\documentclass[a4paper,fleqn]{cas-sc}
\usepackage{adjustbox,lipsum}
\usepackage{multicol}
\usepackage{booktabs}
\usepackage{algorithm}
\usepackage{algpseudocode}
\usepackage{multirow}
\usepackage{amsmath,amssymb,amsfonts}
\usepackage{amsthm}
\usepackage{mathrsfs}
\usepackage{mathtools}
\usepackage{dsfont}
\usepackage{url}
\usepackage{rotating}  
\usepackage{pdflscape}
\usepackage[numbers,sort]{natbib}
\usepackage{hyperref}
\usepackage{float}

\def\tsc#1{\csdef{#1}{\textsc{\lowercase{#1}}\xspace}}
\tsc{WGM}
\tsc{QE}
\tsc{EP}
\tsc{PMS}
\tsc{BEC}
\tsc{DE}

\begin{document}
\let\WriteBookmarks\relax
\def\floatpagepagefraction{1}
\def\textpagefraction{.001}

\shorttitle{GADGET Path Planning Framework}

\shortauthors{MG.Tamizi et~al.}


\title [mode = title]{A Cross-Environment and Cross-Embodiment Path Planning Framework via a Conditional Diffusion Model}   



%
\author[1]{Mehran Ghafarian Tamizi}[style=chinese]



\ead{mehranght@uvic.ca}


\credit{Theoretical development, programming implementation, experimental design, visualization, data curation, conceptualization, and writing the original draft}

\affiliation[1]{organization={Electrical and Computer Engineering Department, University of Victoria},
    addressline={3800 Finnerty Road}, 
    city={Victoria},
    state={BC},
    country={Canada}}

\author[2,3]{Homayoun Honari}[style=chinese]

\ead{homayoun.honari@mila.quebec}

\credit{Research development, manuscript drafting, experimental design, and programming support}

\affiliation[2]{organization={Mila-Quebec AI Institute},
    addressline={6666 Rue Saint-Urbain}, 
    city={Montréal},
    state={QC},
    country={Canada}}

\author[3]{Amir Mehdi Soufi Enayati}[style=chinese]

\ead{amsoufi@uvic.ca}

\credit{Research development, manuscript drafting}

\affiliation[3]{organization={Mechanical Engineering Department, University of Victoria},
    addressline={3800 Finnerty Road}, 
    city={Victoria},
    state={BC},
    country={Canada}}


\author[4]{ Aleksey Nozdryn-Plotnicki}[]

\ead{aleksey@apera.ai}

\credit{Manuscript drafting, review, and editing}

\affiliation[4]{organization={Apera AI},
    addressline={134 Abbott St 501}, 
    city={Vancouver},
    state={BC},
    country={Canada}}

\author[1,3]{Homayoun Najjaran}[style=chinese]

\cormark[1]
\ead{najjaran@uvic.ca}
\cortext[cor1]{Corresponding author}
\credit{Review and
editing, supervision, and funding acquisition}


\begin{abstract}
Path planning for a robotic system in high-dimensional cluttered environments needs to be efficient, safe, and adaptable for different environments and hardware. Conventional methods face high computation time and require extensive parameter tuning, while prior learning-based methods still fail to generalize effectively. The primary goal of this research is to develop a path planning framework capable of generalizing to unseen environments and new robotic manipulators without the need for retraining. We present GADGET (Generalizable and Adaptive Diffusion-Guided Environment-aware Trajectory generation), a diffusion-based planning model that generates joint-space trajectories conditioned on voxelized scene representations as well as start and goal configurations. A key innovation is GADGET’s hybrid dual-conditioning mechanism that combines classifier-free guidance via learned scene encoding with classifier-guided Control Barrier Function (CBF) safety shaping, integrating environment awareness with real-time collision avoidance directly in the denoising process. This design supports zero-shot transfer to new environments and robotic embodiments without retraining. Experimental results show that GADGET achieves high success rates with low collision intensity in spherical-obstacle, bin-picking, and shelf environments, with CBF guidance further improving safety. Moreover, comparative evaluations indicate strong performance relative to both sampling-based and learning-based baselines. Furthermore, GADGET provides transferability across Franka Panda, Kinova Gen3 (6/7-DoF), and UR5 robots, and physical execution on a Kinova Gen3 demonstrates its ability to generate safe, collision-free trajectories in real-world settings. 
\end{abstract}




\maketitle

\section{Introduction}

Path planning is one of the major challenges in robotics, and finding collision-free paths in real time within a complex environment  may yield important industrial applications~\cite{jahanshahi2018robot}. An ideal path-planning framework should find an optimal or near-optimal solution when one exists, operate quickly in real time, and remain robust to changes in the environment. Conventional path planners in industrial settings (where manipulator arms must navigate crowded workspaces) often suffer from long computation times and inefficiency~\cite{tamizi2023review}. In such settings, a planner must generate safe collision-free paths, respond in real time, and maintain high path quality and success rates to meet the demands of safety-critical tasks. Achieving all of these criteria simultaneously is difficult with traditional planners alone. Sampling-based algorithms can find feasible paths but may be too slow or become trapped in local minima in cluttered, high-dimensional environments. Optimization-based methods can produce smooth trajectories, but often require lengthy computations or careful scenario-specific tuning~\cite{weingartshofer2023optimization}. Moreover, industrial applications are increasingly characterized by dynamic and variable conditions, such as changing obstacle configurations, varying task layouts, and the deployment of multiple robot models along a production line~\cite{luo2023survey}.

Given these issues, generalizability has emerged as a key challenge. Path-planning frameworks must be real time and safe, but they must also work in new environments and across different robotic platforms without retraining. Recent learning-based planners can learn from data and produce paths much faster than classical methods, but they often have a limited scope~\cite{tamizi2023review}. Models trained for one workspace or robot arm frequently fail when the scenario changes, necessitating costly retraining for each new setting. These challenges motivate a path-planning approach that combines real-time performance and safety with broad adaptability across tasks, environments, and hardware. The primary goal of this work is to create an adaptable path planning system that can find a safe solution in real-time for different robot tasks. Specifically, the purpose is to have a planning framework that can be used in different robotic arms and environments without retraining for each new scenario. This framework should adapt to changes in the kinematics of the robot or the layout of the environment and produce collision-free paths that respect task constraints and safety requirements.

This paper proposes GADGET which stands for Generalizable and Adaptive Diffusion-Guided Environment-aware Trajectory generation, a diffusion model path planner that leverages recent advances in conditional denoising for trajectory generation. 
GADGET employs both features of classifier-free and classifier-guided diffusion. The workspace is represented by the coordinates of occupied voxels which condition the diffusion model along with the start and goal states, enabling it to generate feasible paths that adapt to new environments and configurations. Moreover, it integrates a Control Barrier Function (CBF) inspired guidance mechanism into the diffusion sampling process. 

The main contributions of this work are summarized as follows:
\begin{itemize}
    \item We propose GADGET (Generalizable and Adaptive Diffusion-Guided Environment-aware Trajectory generation), a conditional diffusion model that generates feasible joint-space trajectories by conditioning on voxelized scene representations and task specifications (start and goal configurations). GADGET achieves zero-shot generalization to both unseen workspace layouts and different manipulator embodiments.
    \item GADGET employs a dual-conditioning strategy that integrates (i) classifier-free guidance for scene and goal awareness through learned voxel embeddings, and (ii) CBF-inspired classifier-guided diffusion for real-time collision avoidance. This hybrid approach balances goal-directed planning with geometric safety enforcement directly within the iterative denoising process.
    \item GADGET achieves cross-embodiment transfer by learning joint-space trajectory priors on a representative manipulator (Franka Panda) and adapting them at inference time through CBF-based safety shaping that incorporates the forward kinematics of the deployed robot. While the diffusion model implicitly encodes kinematic structure from training data, the CBF guidance mechanism provides morphology-specific geometric feedback, enabling zero-shot transfer to manipulators with similar kinematic topologies (Kinova Gen3, UR5) without retraining.
    \item Extensive experiments demonstrate that GADGET achieves high success rates and low collision intensity across diverse environments including spherical obstacles, bin-picking, and shelf manipulation tasks. Real-world execution on a Kinova Gen3 robot further validates the framework's ability to generate safe, collision-free trajectories from sensory voxel reconstructions.
\end{itemize}

The remainder of this paper is organized as follows. Section~\ref{lit} reviews related work and highlights their advantages and disadvantages. Section~\ref{pre} defines the basic concepts needed to better understand this study. Section~\ref{method} introduces the GADGET framework and provides details of the proposed method. Section~\ref{exp} presents the experimental results and compares GADGET with the latest methods. Section~\ref{real} describes the practical implementation of GADGET on Kinova Gen3, and finally Section~\ref{con} concludes the paper.

\section{Related Work} \label{lit}

Robotic arms, which operate across diverse domains, ranging from industrial automation to medical applications~\cite{boscariol2022path}, represent one of the most widely utilized robotic systems in industry. One of the main challenges in robotics is finding collision-free paths in real time in complex environments. According to the literature, path planning methods can be categorized into two main methods; classical and learning-based approaches~\cite{tamizi2023review}.  Potential field approaches~\cite{long2020virtual, conkur2005path}, bio-inspired heuristics~\cite{tian2004effective}, and sampling-based planners~\cite{elbanhawi2014sampling} are among the most widely classical-based planners.

Sampling-based methods is the most notable conventional planners due to their scalability in high-dimensional spaces. Probabilistic Roadmaps (PRM)~\cite{latombe1998probabilistic} is used for multi-query planning by building global roadmaps; however, their high computational cost and suboptimal solutions limit their use in manipulation tasks~\cite{rodriguez2014planning,rosell2011autonomous,karaman2011sampling,ellekilde2013motion}. Rapidly-exploring Random Trees (RRT)~\cite{lavalle1998rapidly} attempt to solve the path planning problem by incrementally exploring configuration spaces. Additionally, RRT is probabilistically complete and suitable or high-dimensional, non-convex problems~\cite{rybus2015application,wang2020collision,rybus2020point}. The main drawback of RRT is that, although it guarantees probabilistic completeness, it often produces suboptimal solutions. To address this limitation,
variants such as Bi-RRT~\cite{kuffner2000efficient,riedlinger2022concept} and RRT*~\cite{karaman2011sampling} improve performance and optimality, while extensions like BIT*~\cite{gammell2015batch} and Informed-RRT~\cite{gammell2014informed} introduce informed search strategies. However, the reliance on dense sampling and expensive collision checking results lang planning time and slow convergence, making these methods unsuitable for real-time robotic manipulation in cluttered environments~\cite{iversen2017benchmarking}.

To mitigate these limitations, learning-based methods have emerged as a powerful alternative. 
hese techniques aim to improve efficiency by leveraging data to bias the search process or replace it entirely with learned policies. Neural samplers~\cite{qureshi2018deeply} accelerate sampling-based planners, while reinforcement learning (RL)~\cite{aleo2010sarsa,duguleana2012obstacle,chang2021reinforcement,li2020motion} enables agents to learn policies for unknown environments. However, RL methods often suffer from data inefficiency, instability due to sparse rewards~\cite{nair2018overcoming}, and the need for extensive computational resources~\cite{xie2019deep,peng2022deep}. Imitation learning provides an alternative by exploiting expert demonstrations. For instance, in~\cite{rahmatizadeh2016learning}, recurrent neural networks were trained for manipulation tasks. Recently, neural planners are used as end-to-end alternatives to classical approaches. Bency et al.~\cite{bency2019neural} proposed a recurrent model for iterative trajectory generation, while MPNet~\cite{qureshi2019motion,qureshi2020motion} demonstrated real-time performance in cluttered environments. Moreover, Tamizi et al.~\cite{tamizi2024end} proposed an end-to-end framework for path planning and collision checking that employs two complementary neural networks: one for predicting subsequent waypoints and another for validating their collision status. Although these methods demonstrate a promising performance, they struggle with adapting to novel configurations and maintaining safety in unseen environments. 

Generative models represent a new paradigm in learning-based planning. These methods learn distributions over feasible trajectories instead of computing paths via explicit optimization or sampling. Generative Adversarial Networks (GANs)-based planners~\cite{goodfellow2020generative,zhang2025generative} employ the generator–discriminator framework to learn feasible trajectory distributions from demonstrations. The discriminator enforces trajectory validity, allowing generated motions to generalize to novel tasks. While GANs offer diversity, their training instability such as mode collapse limits robustness~\cite{zhang2025generative}. Moreover, Variational Autoencoders (VAEs)~\cite{kingma2013auto} learn the latent representations of feasible paths, enabling efficient trajectory synthesis. For example, Hung et al.~\cite{hung2022reaching} perform a latent space optimization for raching. Additionally, in~\cite{ichter2019robot} proposed an embedding-based planner that integrate RRT in latent spaces. VAE-based methods can also capture entire families of homotopically distinct paths~\cite{osa2022motion}, offering flexibility in trajectory generation.

Denoising diffusion probabilistic models (DDPMs)~\cite{ho2020denoising} have recently been adopted in the context of robot motion planning where the whole path was considered as a single data sample going through iterative denoising. These models are trained on datasets of collision-free joint-space trajectories and conditioned on task-relevant inputs such as start and goal states or scene encodings~\cite{carvalho2023motion}. Janner et al.~\cite{janner2022planning} developed Diffuser, where instead of using traditional optimization, full trajectories are directly sampled from a diffusion model. Chi et al.~\cite{chi2023diffusion} conditioned the diffusion process on visual observations, enabling step-by-step policy generation. Ze et al.~\cite{ze20243d} proposed DP3, which leverages point cloud inputs to generate dexterous manipulation trajectories. Huang et al.~\cite{huang2023diffusion} presented SceneDiffuser, embedding physical constraints into the denoising process to unify trajectory generation, optimization, and planning in 3D environments. Across these works, diffusion models have shown the ability to generate diverse, feasible trajectories while generalizing to new goals and scene configurations.

\section{Preliminaries} \label{pre}

In this section, before presenting the proposed framework, the necessary background and notation used throughout this paper will be discussed. This includes a formal definition of the path planning problem for a robotic arm, key concepts related to diffusion models, and a background in CBFs which are later used to guide the diffusion process during trajectory generation.

\subsection{Path Planning}

Path planning can be performed either in the joint space or in the Cartesian (task) space of a robotic arm. In the joint space, path planning refers to the process of computing a continuous and feasible trajectory for the robot’s joints that moves the manipulator from an initial configuration to a desired goal configuration while satisfying kinematic, dynamic, and environmental constraints. Unlike task-space planning, which is based on the end-effector path in workspace coordinates, joint-space planning explicitly defines the path in terms of the robot’s joint angels. 

Let a robotic manipulator be described by a vector of joint variables
\begin{equation}
\mathbf{q} = [q_1, q_2, \dots, q_n]^T \in \mathcal{C},
\end{equation}
where $\mathcal{C} \subseteq \mathbb{R}^n$ is the configuration space (C-space) of the robot with $n$ degrees of freedom. The path planning problem in joint space is to find a continuous mapping $\tau: [0,1] \to \mathcal{C}_{\text{free}}$
such that $\tau(0) = \mathbf{q}_{\text{start}}$, $\tau(1) = \mathbf{q}_{\text{goal}}$, and $\tau(s) \in \mathcal{C}_{\text{free}}$ for all $s \in [0,1]$, where $\mathcal{C}_{\text{free}} \subseteq \mathcal{C}$ denotes the collision-free subset of the configuration space.
The main objective in this formulation is to ensure that every intermediate configuration along the path is both kinematically valid (within joint limits) and free of collisions with the environment or self-collisions.

\subsection{Diffusion Models}

\begin{figure}[t]
\centering
\includegraphics[width=0.8\columnwidth]{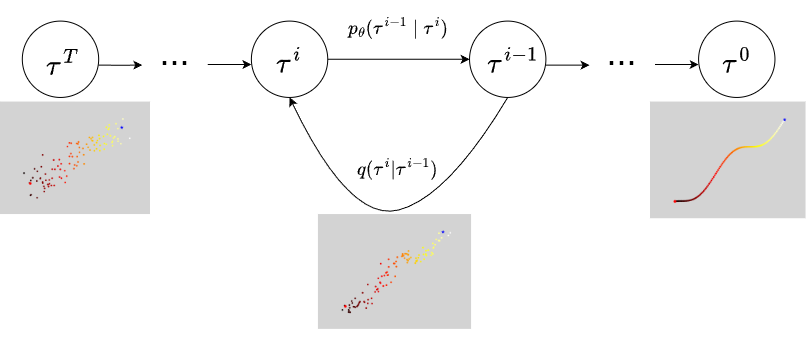}
\caption{Forward and reverse (denoising) process of Diffusion model.}
\label{denoise}
\end{figure}

As a powerful class of generative models, diffusion models offer notable advantages for modeling trajectories and control policies in robotics. DDPMs~\cite{ho2020denoising} as latent variable models synthesize data by using a learned, iterative denoising process. The core of DDPMs includes a a forward diffusion process in which structured data, such as a trajectory, is progressively corrupted by Gaussian noise over multiple time steps until it becomes indistinguishable from pure noise. Then, a neural network is trained to reverse this process by denoising the data step by step in order to reconstruct samples from the original data distribution. During inference, new trajectories are generated by starting from pure noise and applying the learned denoising procedure, producing smooth and feasible path plans (see Figure~\ref{denoise}).

The forward diffusion process is defined by a sequence of conditional Gaussian distributions:

\begin{equation}
    q(\tau^i | \tau^{i-1}) = \mathcal{N} \left( \tau^i ; \sqrt{1 - \beta_i} \, \tau^{i-1}, \, \beta_i \mathbf{I} \right)
\end{equation}

In the above equation:

\begin{itemize}
\item  $q(\tau^i | \tau^{i-1})$ represents the forward diffusion process, i.e., the conditional probability of obtaining the noisy trajectory $\tau^i$ at step $i$, given the trajectory $\tau^{i-1}$ from the previous step.
\item $\tau^i$ is the trajectory at diffusion timestep $i$, after adding noise.
\item $\tau^{i-1}$ is the trajectory at the previous diffusion timestep $i-1$, before further noise is added.
\item $\mathcal{N} ( . ; \mu, \Sigma)$ denotes a Gaussian distribution with mean $\mu$ and covariance $\Sigma$.
\item $\sqrt{1 - \beta_i}$ is a scaling factor applied to $\tau^{i-1}$ which ensures that the variance of the process increases gradually over time.
\item $\beta_i$ is a hyperparameter that controls the amount of noise added at each timestep $i$.
\item $\mathbf{I}$ is the identity matrix, used to define an isotropic Gaussian noise with variance $\beta_i$ in all dimensions.
\end{itemize}

The following equation defines the reverse diffusion process, which is modeled as a sequence of conditional probability distributions that iteratively denoise the trajectory.

\begin{equation}
    p_{\theta}(\tau^{i-1} \mid \tau^i) = \mathcal{N} \left( \tau^{i-1}; \mu_{\theta}(\tau^i, i), \Sigma_{\theta}(\tau^i, i) \right)
\end{equation}
where a neural network parameterized by $\theta$ predicts the mean and variance for sampling $\tau^{i-1}$ from $\tau^i$. The goal here is that to train this network in a way that the original data distribution be reconstructed accurately by the reverse diffusion process with progressively denoising the input. As a result, the loss function used to train the diffusion model can be formulated as follows. This is the simplified training objective commonly used in DDPMs, where the model learns to predict the noise $\epsilon$ added to the original data $\tau^0$ at timestep $i$~\cite{ho2020denoising}.

\begin{equation}
    L(\theta) = \mathbb{E}_{i, \tau^0, \epsilon} \left[ \left\| \epsilon - \epsilon_{\theta} \left( \sqrt{\bar{\alpha}_i} \, \tau^0 + \sqrt{1 - \bar{\alpha}_i} \, \epsilon, \, i \right) \right\|^2 \right]
\end{equation}
where $\epsilon$ is Gaussian noise, $\epsilon_\theta$ is the neural network which predict the noise, and $\bar{\alpha}_i =  \prod_{j=1}^{i} (1 - \beta_j)$ is the cumulative product of the noise schedule. It is important to note that this work adopts the cosine noise schedule proposed by Nichol and Dhariwal~\cite{nichol2021improved}, and assumes a fixed, precomputed variance for the reverse process, i.e., $\Sigma_{\theta}(\tau^i, i) = \Sigma^i$. 

Standard diffusion formulations are often unconditional, which limits their ability to enforce task-specific requirements such as goal reaching, obstacle avoidance, or safety constraints. To address this, guided diffusion techniques introduce additional signals during the reverse denoising process. Classifier-guided diffusion uses an external classifier $C(\cdot \mid y)$ to bias generated trajectories toward desired outcomes. The guided mean at timestep $i$ is given by~\cite{ho2022classifier}:
\begin{equation}
\mu_{\text{guided}}(\tau^i, i) = \mu_{\theta}(\tau^i, i) + \lambda \cdot \Sigma^i \nabla_{\tau^i} \log C(y \mid \tau^i),
\end{equation}
where $\mu_{\theta}$ and $\Sigma^i$ are the mean and variance predicted by the diffusion model, $C(y \mid \tau^i)$ estimates the likelihood of satisfying condition $y$, and $\lambda$ controls guidance strength. In contrast, classifier-free guidance integrates conditioning directly into the generative model, combining conditional and unconditional predictions at inference time~\cite{ho2022classifier}:
\begin{equation}\label{classifier_free}
\epsilon_{\text{guided}} = (1 + \lambda) \cdot \epsilon_{\theta}(\tau^i, i, y) - \lambda \cdot \epsilon_{\theta}(\tau^i, i, \emptyset),
\end{equation}
where $\epsilon_{\theta}(\tau^i, i, y)$ and $\epsilon_{\theta}(\tau^i, i, \emptyset)$ denote conditional and unconditional noise estimates, respectively.

Multi-modality is a key characteristic that makes diffusion models particularly well-suited for path planning tasks. Diffusion models naturally capture the multi-modal nature of planning problems, where multiple feasible paths
can exist for the same task. In addition, diffusion models are able to generate trajectories at the sequence level, ensuring long-horizon coherence and reducing error accumulation. Moreover, the flexible conditioning mechanism allows diffusion-based planners to generalize across novel environments and start-goal pairs. By modeling trajectories as structured sequences, diffusion models also capture temporal dependencies, recombining sub-behaviors into globally consistent path plans~\cite{janner2022planning}. These properties make diffusion-based planning a compelling framework for achieving safe, flexible, and generalizable robotic path generation.

\subsection{Control Barrier Functions}

CBFs~\cite{ames2019control} provide a formal mechanism for enforcing safety constraints in control systems by ensuring forward invariance of a designated safe set. Let us consider a control-affine nonlinear dynamical system of the form:

\begin{equation}
\dot{x} = f(x) + g(x) u
\end{equation}
where \( x \in \mathbb{R}^n \) is the system state, \( u \in \mathbb{R}^m \) is the control input, \( f : \mathbb{R}^n \rightarrow \mathbb{R}^n \) and \( g : \mathbb{R}^n \rightarrow \mathbb{R}^{n \times m} \) are locally Lipschitz continuous functions representing the system dynamics.
\newpage
To encode safety, a continuously differentiable scalar function is defined \( h : \mathbb{R}^n \rightarrow \mathbb{R} \), known as the control barrier function, which characterizes the forward-invariant safe set:

\begin{equation}
\mathcal{C} = \{ x \in \mathbb{R}^n \mid h(x) \geq 0 \}
\end{equation}

The goal is to ensure that if the system starts within \( \mathcal{C} \), it remains in \( \mathcal{C} \) for all future time: \( \mathcal{C} \) is forward invariant. This is guaranteed if the time derivative of \( h(x) \) along the system trajectories satisfies the following inequality:

\begin{equation} \label{eq:cbf_condition}
\sup_{u \in \mathbb{R}^m} \left[ \nabla h(x)^\top (f(x) + g(x) u) \right] \geq -\alpha(h(x))
\end{equation}
for a chosen extended class-\( \mathcal{K} \) function \( \alpha \), which is typically chosen as a linear function \( \alpha(h) = \alpha h \) for some \( \alpha > 0 \). This inequality imposes a constraint on the allowable control inputs \( u \) such that the system does not violate the safety condition \( h(x) \geq 0 \).

Equation~\eqref{eq:cbf_condition} defines a set of control inputs \( \mathcal{K}_{\text{cbf}}(x) \subseteq \mathbb{R}^m \) that are admissible for maintaining forward invariance of \( \mathcal{C} \). CBFs can be integrated into control and optimization frameworks such as quadratic programs (QPs) or reinforcement learning policies to ensure constraint satisfaction in real time~\cite{ames2016control, nguyen2016exponential, agrawal2017discrete}.

In generative path planning, the CBF condition could be utilized in the task of generating trajectories as a differentiable constraint or guidance signal. When applied to diffusion models, this enables incorporating safety-critical knowledge (such as obstacle avoidance) directly into the denoising process without sacrificing the generative flexibility of the model. Using the strengths of diffusion models and CBFs, the following section introduces a path planning framework titled GADGET, Generalizable and Adaptive Diffusion-Guided Environment-aware Trajectory generation, designed for safe, efficient, and generalizable robotic manipulation.




\section{GADGET Framework}
\label{method}

\begin{figure}[t!]
\centering
\includegraphics[width=1\columnwidth]{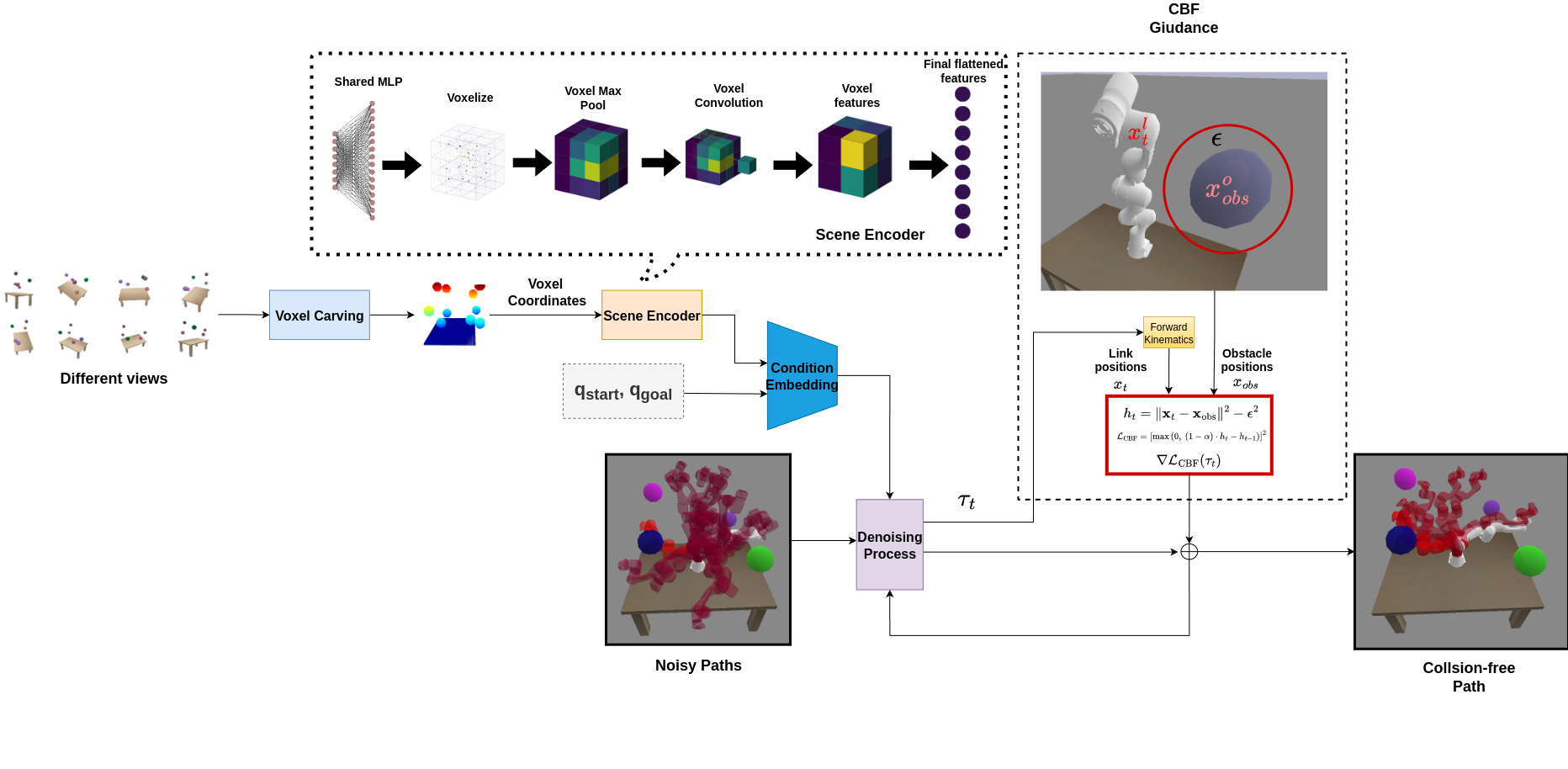}
\caption{Generalizable and Adaptive Diffusion-Guided Environment-aware Trajectory Generation (GADGET) framework.}
\label{gadget}
\end{figure}

The GADGET framework is designed to perform generalizable, collision-free path planning for robotic manipulators across diverse environments. It combines voxel-based scene representation, conditional generative modeling via denoising diffusion, and CBF-inspired safety guidance. The modular design enables transferability across unseen environments and across robot platforms without retraining. Figure~\ref{gadget} provides an overview of the framework.

\subsection{Scene Perception via Voxel Carving}

GADGET employs a voxel carving pipeline to reconstruct the 3D occupancy of the workspace. Multiple depth cameras placed around the robot capture views of the environment. Each depth image is back-projected into 3D space to obtain a point cloud of occupied regions, which is then discretized into a fixed-resolution voxel grid. Voxels along free-space rays between the camera and visible surfaces are marked free. The result is a binary occupancy grid from which the coordinates of occupied voxels are extracted and processed through a scene encoder inspired by~\cite{danielczuk2021object}. Shared MLP layers embed each voxel coordinate, and voxel-wise max pooling aggregates these into a global latent representation $\phi_\psi(s)$. This embedding captures the geometry and spatial distribution of obstacles, providing geometric context to the diffusion model for planning collision-free trajectories.

\subsection{Conditional Embedding}

To guide trajectory generation, the diffusion model is conditioned on both the encoded scene and the robot’s task specification. The condition vector is defined as
$
y = \{ q_s, q_g, \phi_\psi(s) \},
$
where $q_s$ and $q_g$ are the start and goal joint configurations, and $\phi_\psi(s)$ is the latent scene embedding. These elements are fused into a single context vector, enabling the diffusion model to generate trajectories aligned with the environment geometry and task objectives.

\subsection{Diffusion-Based Path Planning with CBF Guidance}
\label{sec:diffusion-based}

GADGET employs a dual-phase guidance methodology that combines task conditioning via classifier-free guidance with CBF-inspired noise perturbations for safety shaping. This method incorporates geometry-aware safety constraints into the denoising process while allowing for a wide variety of condition-driven trajectory generation.

A trajectory denoted by $\tau$ is represented as $\tau \in \mathbb{R}^{T \times d}$, with $T$ being the time horizon and $d$ the configuration space dimension. The trajectory generator, which is a conditional denoising diffusion model, is represented as $\epsilon_\theta(\tau_t, t, y)$ that forecasts the noise to be eliminated from the noisy trajectory $\tau_t$ at time $t$ conditioned on scene/task $y$. The model is trained using expert demonstrations $\tau_0 \sim \mathcal{D}$ by minimizing:
\begin{equation}
\mathcal{L}(\theta,\psi) = 
\mathbb{E}_{\tau_0, t, \epsilon} \left[ 
\| \epsilon - \epsilon_\theta(\tau_t, t, y) \|^2 
\right],
\end{equation}
where $y$ may be randomly masked with probability $p_{\mathrm{drop}}$~\cite{ho2022classifier} to support classifier-free guidance during inference.
\begin{equation}
\epsilon_{\mathrm{guided}}(\tau_t, t, y) = 
(1 + \lambda_{\mathrm{cfg}}) \epsilon_\theta(\tau_t, t, y) 
- \lambda_{\mathrm{cfg}} \epsilon_\theta(\tau_t, t, \emptyset),
\end{equation}
with guidance coefficient $\lambda_{\mathrm{cfg}}\geq 0$. The reverse process mean is then:
\begin{equation}
\mu_t(\tau_t, t, y) = \frac{1}{\sqrt{\alpha_t}} 
\left( 
\tau_t - \frac{1 - \alpha_t}{\sqrt{1 - \bar{\alpha}_t}} 
\, \epsilon_{\mathrm{guided}}(\tau_t, t, y) 
\right),
\end{equation}
where $\alpha_t$ and $\bar{\alpha}_t$ follow a standard diffusion noise schedule.

\vspace{1ex}
\noindent\textbf{Guided Diffusion as Conditional Sampling.}
Sampling is biased toward desirable, safe behaviors by employing a guidance signal $C$:
\begin{equation}
    \tilde p(\tau \mid y) = p(\tau \mid y, C)\propto  p_\theta(\tau \mid y)p(\tau \mid C),
\end{equation}
where $C$ denotes arbitrary inference-time constraints (e.g., safety constraints). The reverse diffusion process is sampled from:
\begin{equation}
p(\tau_0 \mid y, C) = p(\tau_N|y,C)\prod_{t=1}^{N}p(\tau_{t-1} \mid \tau_t,y,C)
\end{equation}
with $p(\tau_N|y,C)$ standard Gaussian, so it suffices to recursively sample
\begin{equation} \label{eq:perturbed-transition}
    p(\tau_{t-1} \mid \tau_t, y, C) \propto p_\theta(\tau_{t-1} \mid \tau_t,y)p(C\mid\tau_{t-1}) 
\end{equation}
where $p_\theta(\tau_{t-1} \mid \tau_t,y)$ is modeled with a classifier-free diffusion model. Since the reverse transition is Gaussian,
\begin{equation}\label{class_free_denoise}
    \log p_\theta(\tau_{t-1} \mid \tau_t,y) \propto -\frac{1}{2}(\tau_{t-1}-\mu_t)^{T}\Sigma_t^{-1}(\tau_{t-1}-\mu_t)
\end{equation}
a first-order Taylor expansion of $\log p(C\mid\tau_{t-1})$ around $\mu_t$ ~\cite{carvalho2023motion}:
\begin{equation}\label{cons_guide}
    \log p(C\mid\tau_{t-1}) \approx \log p(C\mid\mu_t) + (\tau_{t-1}-\mu_t)^\top g
\end{equation}
with $g = \nabla_{\tau}\log p(C|\tau)|_{\tau=\mu_t}$.
Substituting Eq.~\ref{class_free_denoise} and Eq.~\ref{cons_guide} into Eq.~\ref{eq:perturbed-transition} shows that the new reverse transition has mean $\mu_t + \Sigma_t g$.

Assuming independent constraints, the guidance likelihood expresses the probability that a trajectory $\tau$ satisfies all desired constraints as an exponential-family distribution is given by:
\begin{equation}
p(C|\tau)\propto \prod_i \exp(-J_i(\tau)),
\end{equation}
where each $J_i(\tau)$ represents the penalty associated with violating the $i$-th constraint or safety requirement. The resulting guidance direction simplifies to:
\begin{equation}
    g = -\sum_i \nabla_{\tau}J_i(\tau)\vert_{\tau = \mu_t}
\end{equation}

\vspace{1ex}
\noindent\textbf{Discrete-Time Dynamical System and CBF Definition.}  
The denoising process can be viewed as a discrete-time dynamical system with state $\tau_t$ and update rule:
\begin{equation}
\tau_{t-1} = \tau_t + \delta_t,
\end{equation}
where $\delta_t$ denotes the learned denoising direction.

For this discrete system, a CBF $h(\tau)$ is defined to encode safety constraints. Specifically, for each robot link $\ell$ and obstacle $o$, we define the safety function:
\begin{equation}\label{CBF_can}
h_t^{(\ell,o)} = \big\| \mathbf{x}_t^{(\ell)} - \mathbf{x}_{\mathrm{obs}}^{(o)} \big\|^2 - \epsilon^2,
\end{equation}
where $\mathbf{x}_t^{(\ell)} = f_{\mathrm{FK}}^{(\ell)}(\tau_t)$ is the Cartesian position of link $\ell$ at step $t$, obtained via forward kinematics, and $\epsilon > 0$ is the safety margin specifying minimum allowable clearance Moreover, $f_{\mathrm{FK}}^{(\ell)}(\tau_t)$ is a differentiable function.

In classical CBF theory, the time derivative of the safety function satisfies:
\begin{equation}
\dot{h}(\tau) = \nabla h(\tau)^T \dot{\tau} \geq -\alpha h(\tau)
\end{equation}
In our discrete-time setting, where \( \dot{\tau} \approx \delta_t \), a finite-difference approximation is used. As a result, the discrete-time CBF condition requires that safety does not degrade too quickly between steps:
\begin{equation}
h_{t-1}^{(\ell,o)} \geq (1 - \alpha_{\mathrm{cbf}}) h_t^{(\ell,o)},
\label{eq:cbf_discrete}
\end{equation}
where $\alpha_{\mathrm{cbf}}$ is a parameter within the interval (0,1) that controls the amount of conservativeness in the barrier.  This guarantees that the trajectory update $\delta_t$ is still safe by imposing a restriction on $h$ to not decrease beyond a certain rate. Violations of Eq.~\eqref{eq:cbf_discrete} indicate unsafe movements toward obstacles.
In our formulation, \( h_t \) measures the signed squared distance between robot links and obstacles at denoising step \( t \), providing a quantitative measure of safety. Positive values mean that the robot is keeping a minimum distance of \(\epsilon\) from obstacles, while negative values signal violations of the constraints.

\vspace{1ex}
\noindent\textbf{Soft Penalty for CBF Violations.}  
To enforce this constraint softly during sampling, we define a differentiable penalty over all link–obstacle pairs:
\begin{equation}\label{eq:cbf_loss}
\mathcal{L}_{\mathrm{CBF}}(\tau_t) = \frac{1}{L O} \sum_{\ell=1}^{L} \sum_{o=1}^{O} 
\left[ \max \left( 0, (1 - \alpha_{\mathrm{cbf}}) h_t^{(\ell,o)} - h_{t-1}^{(\ell,o)} \right) \right]^2,
\end{equation}
where $L$ and $O$ denote the number of links of the robot and obstacles respectively. Since each $h_t^{(\ell,o)}$ is dependent on the forward kinematics, the gradient $\nabla_{\tau_t} \mathcal{L}_{\mathrm{CBF}}(\tau_t)$ transmits the geometric feedback throughout the kinematic chain to guide the safe denoising process.
 
This CBF penalty function fits naturally into the framework of constraint guidance used in diffusion sampling. Each cost term $J_i(\tau)$ corresponds to an individual link–obstacle penalty:
\begin{equation}
J_i(\tau_t) \equiv \left[ \max \left( 0, (1 - \alpha_{\mathrm{cbf}}) h_t^{(\ell,o)} - h_{t-1}^{(\ell,o)} \right) \right]^2,
\end{equation}
where the index $i$ maps uniquely to $(\ell,o)$. The total penalty is a sum over $i$, 
\[
\mathcal{L}_{\mathrm{CBF}}(\tau_t) = \frac{1}{L O} \sum_{i} J_i(\tau_t),
\]
which guides the diffusion process via gradients of these penalties.

\vspace{1ex}
\noindent\textbf{Final Reverse Update.}  
The overall guided denoising update combines classifier-free guidance with the CBF perturbation:
\begin{equation}\label{eq:final_updat}
\tau_{t-1} = \mu_t(\tau_t, t, y) - \lambda_{\mathrm{cbf}}\, \nabla_{\tau_t} \mathcal{L}_{\mathrm{CBF}}(\tau_t),
\end{equation}
where $\lambda_{\mathrm{cbf}}$ is the parameter that determines the effect of safety shaping. The guidance acts as a soft barrier at each denoising step, preventing the trajectory from entering unsafe areas and at the same time allowing for task compliance.

\begin{algorithm}[t!]
\caption{\textbf{GADGET: Diffusion-Based Path Planning with CBF Guidance}}
\label{alg:gadget}

\textbf{TRAINING}\\
\textbf{Input:} Expert dataset $\mathcal{D}$ (collision-free trajectories), diffusion model $\epsilon_\theta$, noise schedule $\{\bar{\alpha}_t\}_{t=1}^N$, learning rate $\eta$, classifier-free dropout $p_{\mathrm{drop}}$ \\
\textbf{while} training not converged \textbf{do}
\begin{algorithmic}[1]
    \State Sample expert trajectory $\tau_0 \sim \mathcal{D}$, timestep $t \sim \mathcal{U}(1, N)$, noise $\epsilon \sim \mathcal{N}(0, \mathbf{I})$
    \State Forward diffuse trajectory:
    \[
        \tau_t = \sqrt{\bar{\alpha}_t}\,\tau_0 + \sqrt{1 - \bar{\alpha}_t}\,\epsilon
    \]
    \State With probability $p_{\mathrm{drop}}$, set $y = \emptyset$, else $y = \{ q_s, q_g, \phi_\psi(s) \}$
    \State Compute denoising loss:
    \[
        \mathcal{L}(\theta) = \left\| \epsilon - \epsilon_\theta(\tau_t, t, y) \right\|^2
    \]
    \State Update network parameters: $\theta \gets \theta - \eta \nabla_\theta \mathcal{L}(\theta)$
\end{algorithmic}

\vspace{0.7em}
\textbf{INFERENCE}\\
\textbf{Input:} Trained model $\epsilon_\theta$, start/goal $(q_s, q_g)$, safety weight $\lambda_{\mathrm{cbf}}$, safety margin $\epsilon$, temporal factor $\alpha_{\mathrm{cbf}}$, classifier-free weight $\lambda_{\mathrm{cfg}}$ \\
\begin{algorithmic}[1]
    \State Acquire multi-view depth images $\{\mathcal{D}_i\}_{i=1}^M$
    \State Reconstruct occupied voxel coordinates $P = \texttt{VoxelCarve}(\{\mathcal{D}_i\})$ \Comment{$P \in \mathbb{R}^{N \times 3}$}
    \State Encode scene: $\phi_\psi(s) = \texttt{SceneEncoder}(P)$
    \State Initialize noisy trajectory: $\tau_N \sim \mathcal{N}(0, \mathbf{I})$
    \For{$t = N, \dots, 1$}
        \State Compute conditional and unconditional noise:
        \[
            \epsilon_{\mathrm{cond}} = \epsilon_\theta(\tau_t, t, y), \quad 
            \epsilon_{\mathrm{uncond}} = \epsilon_\theta(\tau_t, t, \emptyset)
        \]
        \State Apply classifier-free guidance:
        \[
            \epsilon_{\mathrm{guided}} = (1 + \lambda_{\mathrm{cfg}})\,\epsilon_{\mathrm{cond}} - \lambda_{\mathrm{cfg}}\,\epsilon_{\mathrm{uncond}}
        \]
        \State Predict reverse-process mean:
        \[
            \mu_t = \frac{1}{\sqrt{\alpha_t}} \left( \tau_t - \frac{1 - \alpha_t}{\sqrt{1 - \bar{\alpha}_t}}\,\epsilon_{\mathrm{guided}} \right)
        \]
        \State Compute signed distance using Eq.~\ref{CBF_can}
        \State Compute safety gradient: $\nabla_{\tau_t}\mathcal{L}_{\mathrm{CBF}}(\tau_t)$ using Eq.~\ref{eq:cbf_loss}
        \State Update trajectory with CBF-guided correction:
        \[
            \tau_{t-1} = \mu_t - \lambda_{\mathrm{cbf}} \nabla_{\tau_t}\mathcal{L}_{\mathrm{CBF}}(\tau_t)
        \]
    \EndFor
    \State \textbf{Output:} Collision-free trajectory $\tau_0$
\end{algorithmic}
\end{algorithm}

Algorithm~\ref{alg:gadget} shows the training and inference process of GADGET. The combined use of classifier-free and CBF-based guidance can be interpreted as energy shaping: the diffusion model parameterizes the conditional distribution $p_\theta(\tau_t \mid y)$, and the CBF term reshapes it toward safety-compliant regions of the trajectory space. This unified formulation provides:

\begin{itemize}
    \item \textbf{Multi-Modality and Flexibility:} The stochastic diffusion process samples diverse, feasible trajectories, capturing multiple path hypotheses for a given task.
    \item \textbf{Environment Awareness:} Classifier-free conditioning on $(q_s, q_g, \phi_\psi(s))$ ensures that generated trajectories are well aligned with cluttered and previously unseen environments.
    \item \textbf{Safety Enforcement and Cross-Embodiment Transfer:} The CBF-based loss penalizes proximity to obstacles at every denoising step. Because this safety shaping relies solely on geometric distances computed from robot-specific forward kinematics, the same model transfers seamlessly across different manipulators without retraining.
\end{itemize}

\section{Experimental Results} \label{exp}

To evaluate the proposed framework, we investigate the effectiveness of GADGET through the following key questions:

\begin{itemize}
    \item \textbf{Q1} Can it generalize to unseen environments without retraining?
    \item \textbf{Q2} Can it transfer to different robot arms with distinct kinematics?
    \item \textbf{Q3} Does CBF-guided diffusion improve safety in different environments?
    \item \textbf{Q4} How does GADGET compare to state-of-the-art learning-based and sampling-based planners?
\end{itemize}

\begin{figure}[t]
\centering
\includegraphics[width=0.8\columnwidth, trim=150 30 100 0, clip]{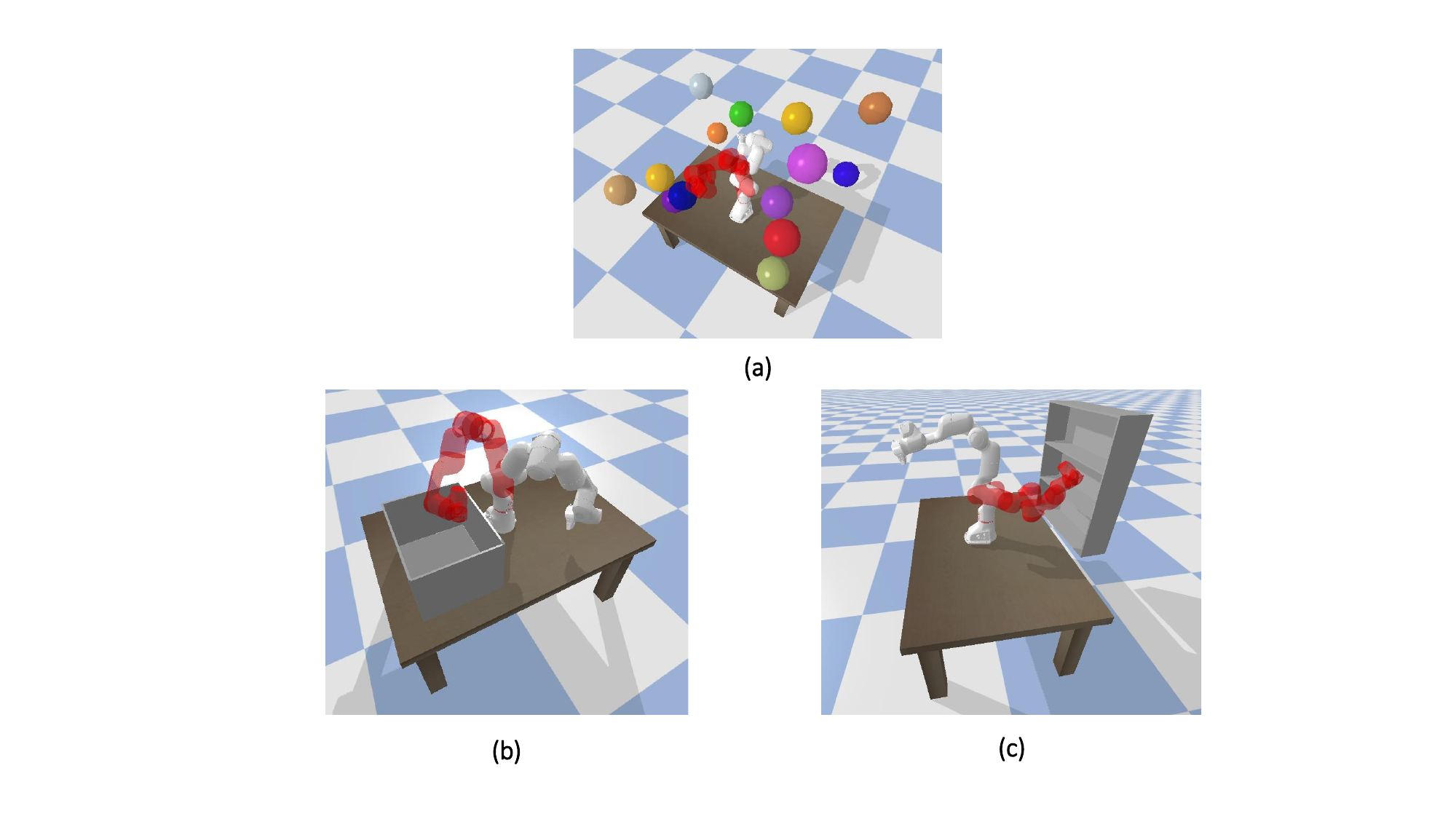}
\caption{Simulation environments. (a) Spherical Obstacles: Used for training; during training and testing, obstacle positions are randomized.
(b) Bin Picking: A test-only environment not encountered during training.
(c) Shelf Manipulation: Another test-only environment unseen during training, representing structured manipulation tasks.
In all scenes, the red shadow of the robot denotes the target (goal) configuration.}
\label{py_envs}
\end{figure}

To answer these questions, GADGET is evaluated across a variety of simulated scenarios designed to test its generalization, safety, and planning performance. The training phase includes
\textbf{Environments:} We generate 10,000 randomized training scenes populated with 10–16 spherical obstacles placed in the robot's workspace (Figure~\ref{py_envs}).
\textbf{Expert Planner:} Ground-truth collision-free trajectories are computed using the BiRRT algorithm.
\textbf{Robot:} A 7-DoF Franka Emika Panda arm is used for data collection and trajectory supervision.

In order to evaluate the generalization capability of GADGET, it is compared against different sampling-based and learning-based path planners. Each of these methods chosen for their strengths and widespread adoption in the field.

\textbf{Bi-RRT} is a classical sampling-based planner that incrementally grows two trees from the start and goal configurations, attempting to connect them. It is probabilistically complete and is often employed as an expert oracle; however, it typically produces suboptimal paths, especially in cluttered environments.

\textbf{BIT*} (Batch Informed Trees)~\cite{gammell2015batch} is an asymptotically optimal sampling-based planner that employs the efficiency of informed graph search like A* and the flexibility of incremental sampling. Although it generated near-optimal paths, it can become computationally expensive, often leading to large planning times in real-time settings.

\textbf{MPNet}~\cite{qureshi2020motion} is one of the first learning-based planners which employs a point cloud encoder together with a neural planner to directly predict a feasible trajectory.

\textbf{MPD} (Motion Planning Diffusion)~\cite{carvalho2023motion} is a recent diffusion-based method which learns priors over robot trajectories. Although it is effective in a constant scene, since it does not explicitly encode scene geometry and just using collision cost as guiding during inference, its performance degrades significantly when new environment.

\textbf{DP3}~\cite{ze20243d} is an imitation learning method which integrates point cloud representations with diffusion policies to generate robot actions.

To quantitatively compare GADGET against baseline planners, the following metrics is used:
\begin{itemize}
    \item \textbf{Success Rate (SR)}: The percentage of planning queries for which at least one generated path is entirely collision free and reaches the goal configuration within a fixed distance. In our experiments, for each start-goal query, we sample 30 trajectories, and the query is considered successful if at least one of them satisfies the criteria. This definition is based on~\cite{carvalho2023motion}, and reflects the multi-modal nature of diffusion models, which can generate diverse candidate solutions for the same query.
    
    \item \textbf{Collision Intensity (CI)}: The percentage of the waypoints that are in collision with obstacles. This metric reflects the safety of the generated trajectories. 
    
    \item \textbf{Planning Time (T)}: The amount of time (in seconds) required to generate a collision-free path for a random start-goal query. In case of the diffusion-based planners, 30 trajectories sampled for each query and collision checking is is performed sequentially. The planning time is recorded at the point when the first collision-free trajectory is found.

    \item \textbf{Path Length (PL):} The length of the generated paths (in radians) which is measured as the sum of Euclidean distances between consecutive waypoints in joint space.

\end{itemize}

\begin{table}[t]
\centering
\caption{Performance of GADGET and baseline planners across diverse test environments.}
\vspace{1em}
\label{benchmarking}
\begin{adjustbox}{max width=\columnwidth}
\begin{tabular}{c|cccc|cccc|cccc}
\hline
                                                                                                    & \multicolumn{4}{c|}{}                                                                                                                                                                                                              & \multicolumn{4}{c|}{}                                                                                                                                                                                                              & \multicolumn{4}{c}{}                                                                                                                                                                                                               \\
                                                                                                    & \multicolumn{4}{c|}{\multirow{-2}{*}{\textbf{Spherical Obs}}}                                                                                                                                                                      & \multicolumn{4}{c|}{\multirow{-2}{*}{\textbf{Bin Picking}}}                                                                                                                                                                        & \multicolumn{4}{c}{\multirow{-2}{*}{\textbf{Shelf Manipulation}}}                                                                                                                                                                  \\ \cline{2-13} 
                                                                                                    &                                                         &                                                        &                                                       &                                                         &                                                         &                                                        &                                                       &                                                         &                                                         &                                                        &                                                       &                                                         \\
\multirow{-4}{*}{Method}                                                                            & \multirow{-2}{*}{T (s)}                                 & \multirow{-2}{*}{CI (\%)}                              & \multirow{-2}{*}{PL (rad)}                            & \multirow{-2}{*}{SR (\%)}                               & \multirow{-2}{*}{T (s)}                                 & \multirow{-2}{*}{CI (\%)}                              & \multirow{-2}{*}{PL (rad)}                            & \multirow{-2}{*}{SR (\%)}                               & \multirow{-2}{*}{T (s)}                                 & \multirow{-2}{*}{CI (\%)}                              & \multirow{-2}{*}{PL (rad)}                            & \multirow{-2}{*}{SR (\%)}                               \\ \hline
Bi-RRT                                                                                              & 0.24$\pm$0.12                                           & \_                                                     & 11.1$\pm$4.4                                          & 99.9                                                    & 0.22$\pm$ 0.14                                          & \_                                                     & 10.7$\pm$3.9                                          & 100                                                     & 0.37$\pm$0.15                                           & \_                                                     & 10.7$\pm$4.1                                          & 100                                                     \\
                                                                                                    &                                                         &                                                        &                                                       &                                                         &                                                         &                                                        &                                                       &                                                         &                                                         &                                                        &                                                       &                                                         \\
\multirow{-2}{*}{BIT*}                                                                              & \multirow{-2}{*}{1.02$\pm$0.02}                         & \multirow{-2}{*}{\_}                                   & \multirow{-2}{*}{8.1$\pm$2.6}                         & \multirow{-2}{*}{94.9}                                  & \multirow{-2}{*}{1.01$\pm$0.01}                         & \multirow{-2}{*}{\_}                                   & \multirow{-2}{*}{7.6$\pm$2.1}                         & \multirow{-2}{*}{90.1}                                  & \multirow{-2}{*}{1.01$\pm$0.02}                         & \multirow{-2}{*}{\_}                                   & \multirow{-2}{*}{8.0$\pm$2.3}                         & \multirow{-2}{*}{77.8}                                  \\
                                                                                                    &                                                         &                                                        &                                                       &                                                         &                                                         &                                                        &                                                       &                                                         &                                                         &                                                        &                                                       &                                                         \\
\multirow{-2}{*}{MPNet}                                                                             & \multirow{-2}{*}{0.22$\pm$0.14}                         & \multirow{-2}{*}{44.1}                                 & \multirow{-2}{*}{7.9$\pm$1.7}                         & \multirow{-2}{*}{53.8}                                  & \multirow{-2}{*}{0.26$\pm$0.15}                         & \multirow{-2}{*}{12.9}                                 & \multirow{-2}{*}{7.6$\pm$1.7}                         & \multirow{-2}{*}{38.7}                                  & \multirow{-2}{*}{0.31$\pm$0.20}                         & \multirow{-2}{*}{22.3}                                 & \multirow{-2}{*}{8.1$\pm$1.9}                         & \multirow{-2}{*}{35.1}                                  \\
                                                                                                    &                                                         &                                                        &                                                       &                                                         &                                                         &                                                        &                                                       &                                                         &                                                         &                                                        &                                                       &                                                         \\
\multirow{-2}{*}{MPD}                                                                               & \multirow{-2}{*}{1.15$\pm$0.04}                         & \multirow{-2}{*}{10.3}                                 & \multirow{-2}{*}{7.2$\pm$1.5}                         & \multirow{-2}{*}{90.0}                                  & \multirow{-2}{*}{1.14$\pm$0.04}                         & \multirow{-2}{*}{14.5}                                 & \multirow{-2}{*}{7.6$\pm$1.6}                         & \multirow{-2}{*}{52.8}                                  & \multirow{-2}{*}{1.14$\pm$0.09}                         & \multirow{-2}{*}{20.5}                                 & \multirow{-2}{*}{7.8$\pm$1.5}                         & \multirow{-2}{*}{26.2}                                  \\
                                                                                                    &                                                         &                                                        &                                                       &                                                         &                                                         &                                                        &                                                       &                                                         &                                                         &                                                        &                                                       &                                                         \\
\multirow{-2}{*}{DP3}                                                                               & \multirow{-2}{*}{0.42$\pm$0.11}                         & \multirow{-2}{*}{14.6}                                 & \multirow{-2}{*}{8.4$\pm$2.3}                         & \multirow{-2}{*}{79.4}                                  & \multirow{-2}{*}{0.41$\pm$0.13}                         & \multirow{-2}{*}{5.4}                                  & \multirow{-2}{*}{8.2$\pm$2.3}                         & \multirow{-2}{*}{76.3}                                  & \multirow{-2}{*}{0.48$\pm$0.19}                         & \multirow{-2}{*}{6.9}                                  & \multirow{-2}{*}{8.2$\pm$2.3}                         & \multirow{-2}{*}{67.8}                                  \\
\cellcolor[HTML]{c0c0c0}                                                                            & \cellcolor[HTML]{c0c0c0}                                & \cellcolor[HTML]{c0c0c0}                               & \cellcolor[HTML]{c0c0c0}                              & \cellcolor[HTML]{c0c0c0}                                & \cellcolor[HTML]{c0c0c0}                                & \cellcolor[HTML]{c0c0c0}                               & \cellcolor[HTML]{c0c0c0}                              & \cellcolor[HTML]{c0c0c0}                                & \cellcolor[HTML]{c0c0c0}                                & \cellcolor[HTML]{c0c0c0}                               & \cellcolor[HTML]{c0c0c0}                              & \cellcolor[HTML]{c0c0c0}                                \\
\multirow{-2}{*}{\cellcolor[HTML]{c0c0c0}\begin{tabular}[c]{@{}c@{}}GADGET\\ w.o CBF\end{tabular}}  & \multirow{-2}{*}{\cellcolor[HTML]{c0c0c0}0.43$\pm$0.09} & \multirow{-2}{*}{\cellcolor[HTML]{c0c0c0}13.9}         & \multirow{-2}{*}{\cellcolor[HTML]{c0c0c0}8.1$\pm$2.2} & \multirow{-2}{*}{\cellcolor[HTML]{c0c0c0}79.9}          & \multirow{-2}{*}{\cellcolor[HTML]{c0c0c0}0.45$\pm$0.12} & \multirow{-2}{*}{\cellcolor[HTML]{c0c0c0}5.2}          & \multirow{-2}{*}{\cellcolor[HTML]{c0c0c0}8.5$\pm$2.5} & \multirow{-2}{*}{\cellcolor[HTML]{c0c0c0}85.2}          & \multirow{-2}{*}{\cellcolor[HTML]{c0c0c0}0.62$\pm$0.23} & \multirow{-2}{*}{\cellcolor[HTML]{c0c0c0}6.5}          & \multirow{-2}{*}{\cellcolor[HTML]{c0c0c0}8.4$\pm$2.3} & \multirow{-2}{*}{\cellcolor[HTML]{c0c0c0}74.2}          \\
\cellcolor[HTML]{c0c0c0}                                                                            & \cellcolor[HTML]{c0c0c0}                                & \cellcolor[HTML]{c0c0c0}                               & \cellcolor[HTML]{c0c0c0}                              & \cellcolor[HTML]{c0c0c0}                                & \cellcolor[HTML]{c0c0c0}                                & \cellcolor[HTML]{c0c0c0}                               & \cellcolor[HTML]{c0c0c0}                              & \cellcolor[HTML]{c0c0c0}                                & \cellcolor[HTML]{c0c0c0}                                & \cellcolor[HTML]{c0c0c0}                               & \cellcolor[HTML]{c0c0c0}                              & \cellcolor[HTML]{c0c0c0}                                \\
\multirow{-2}{*}{\cellcolor[HTML]{c0c0c0}\begin{tabular}[c]{@{}c@{}}GADGET\\ with CBF\end{tabular}} & \multirow{-2}{*}{\cellcolor[HTML]{c0c0c0}0.59$\pm$0.08} & \multirow{-2}{*}{\cellcolor[HTML]{c0c0c0}\textbf{8.2}} & \multirow{-2}{*}{\cellcolor[HTML]{c0c0c0}8.3$\pm$2.2} & \multirow{-2}{*}{\cellcolor[HTML]{c0c0c0}\textbf{95.5}} & \multirow{-2}{*}{\cellcolor[HTML]{c0c0c0}0.60$\pm$0.10} & \multirow{-2}{*}{\cellcolor[HTML]{c0c0c0}\textbf{4.6}} & \multirow{-2}{*}{\cellcolor[HTML]{c0c0c0}8.7$\pm$2.3} & \multirow{-2}{*}{\cellcolor[HTML]{c0c0c0}\textbf{92.2}} & \multirow{-2}{*}{\cellcolor[HTML]{c0c0c0}0.77$\pm$0.23} & \multirow{-2}{*}{\cellcolor[HTML]{c0c0c0}\textbf{5.9}} & \multirow{-2}{*}{\cellcolor[HTML]{c0c0c0}8.7$\pm$2.4} & \multirow{-2}{*}{\cellcolor[HTML]{c0c0c0}\textbf{80.1}} \\ \hline
\end{tabular}
\end{adjustbox}
\end{table}

\begin{table}[t]
\centering
\caption{Cross-robot generalization performance of GADGET compared to baselines without retraining.}
\vspace{1em}
\label{tab:robot-transfer}
\begin{adjustbox}{max width=\columnwidth}
\begin{tabular}{c|lccl|lccl|cccl|cccl}
\hline
                                                 & \multicolumn{4}{c|}{}                                                                                                                                                                                                              & \multicolumn{4}{c|}{}                                                                                                                                                                                                              & \multicolumn{4}{c|}{}                                                                                                                                                                                                               & \multicolumn{4}{c}{}                                                                                                                                                                                                       \\
                                                 & \multicolumn{4}{c|}{\multirow{-2}{*}{\textbf{Franka Panda}}}                                                                                                                                                                             & \multicolumn{4}{c|}{\multirow{-2}{*}{\textbf{Gen3 (7DoF)}}}                                                                                                                                                                        & \multicolumn{4}{c|}{\multirow{-2}{*}{\textbf{Gen3 (6DoF)}}}                                                                                                                                                                         & \multicolumn{4}{c}{\multirow{-2}{*}{\textbf{UR5}}}                                                                                                                                                                         \\ \cline{2-17} 
                                                 & \multicolumn{1}{c}{}                                    &                                                        &                                                         &                                                       & \multicolumn{1}{c}{}                                    &                                                        &                                                         &                                                       &                                                         &                                                        &                                                         &                                                        &                                                         &                                                         &                                                &                                                       \\
\multirow{-4}{*}{Method}                         & \multicolumn{1}{c}{\multirow{-2}{*}{T(s)}}              & \multirow{-2}{*}{CI (\%)}                              & \multirow{-2}{*}{SR (\%)}                               & \multirow{-2}{*}{PL (rad)}                            & \multicolumn{1}{c}{\multirow{-2}{*}{T(s)}}              & \multirow{-2}{*}{CI (\%)}                              & \multirow{-2}{*}{SR (\%)}                               & \multirow{-2}{*}{PL (rad)}                            & \multirow{-2}{*}{T(s)}                                  & \multirow{-2}{*}{CI (\%)}                              & \multirow{-2}{*}{SR (\%)}                               & \multirow{-2}{*}{PL (rad)}                             & \multirow{-2}{*}{T(s)}                                  & \multirow{-2}{*}{CI (\%)}                               & \multirow{-2}{*}{SR(\%)}                       & \multirow{-2}{*}{PL (rad)}                            \\ \hline
                                                 &                                                         &                                                        &                                                         &                                                       &                                                         &                                                        &                                                         &                                                       &                                                         &                                                        &                                                         &                                                        &                                                         &                                                         &                                                &                                                       \\
\multirow{-2}{*}{MPD}                            & \multirow{-2}{*}{1.15$\pm$0.10}                         & \multirow{-2}{*}{10.3}                                 & \multirow{-2}{*}{90.0}                                  & \multirow{-2}{*}{7.3$\pm$1.5}                         & \multirow{-2}{*}{1.08$\pm$0.04}                         & \multirow{-2}{*}{17.8}                                 & \multirow{-2}{*}{55.8}                                  & \multirow{-2}{*}{9.8$\pm$1.9}                         & \multirow{-2}{*}{1.01$\pm$0.05}                         & \multirow{-2}{*}{21.2}                                 & \multirow{-2}{*}{46.8}                                  & \multirow{-2}{*}{7.7$\pm$1.6}                          & \multirow{-2}{*}{1.04$\pm$0.03}                         & \multirow{-2}{*}{29.4}                                  & \multirow{-2}{*}{38.9}                         & \multirow{-2}{*}{7.9$\pm$2.7}                         \\
                                                 &                                                         &                                                        &                                                         &                                                       &                                                         &                                                        &                                                         &                                                       &                                                         &                                                        &                                                         &                                                        &                                                         &                                                         &                                                &                                                       \\
\multirow{-2}{*}{DP3}                            & \multirow{-2}{*}{0.42$\pm$0.12}                         & \multirow{-2}{*}{14.6}                                 & \multirow{-2}{*}{79.4}                                  & \multirow{-2}{*}{8.4$\pm$2.3}                         & \multirow{-2}{*}{0.42$\pm$0.11}                         & \multirow{-2}{*}{15.7}                                 & \multirow{-2}{*}{72.6}                                  & \multirow{-2}{*}{9.4$\pm$2.7}                         & \multirow{-2}{*}{0.38$\pm$0.08}                         & \multirow{-2}{*}{14.1}                                 & \multirow{-2}{*}{75.4}                                  & \multirow{-2}{*}{8.6$\pm$2.6}                          & \multirow{-2}{*}{0.39$\pm$0.09}                         & \multirow{-2}{*}{22.4}                                  & \multirow{-2}{*}{52.2}                         & \multirow{-2}{*}{8.2$\pm$2.4}                         \\
\cellcolor[HTML]{c0c0c0}                         & \cellcolor[HTML]{c0c0c0}                                & \cellcolor[HTML]{c0c0c0}                               & \cellcolor[HTML]{c0c0c0}                                & \cellcolor[HTML]{c0c0c0}                              & \cellcolor[HTML]{c0c0c0}                                & \cellcolor[HTML]{c0c0c0}                               & \cellcolor[HTML]{c0c0c0}                                & \cellcolor[HTML]{c0c0c0}                              & \cellcolor[HTML]{c0c0c0}                                & \cellcolor[HTML]{c0c0c0}                               & \cellcolor[HTML]{c0c0c0}                                & \cellcolor[HTML]{c0c0c0}                               & \cellcolor[HTML]{c0c0c0}                                & \cellcolor[HTML]{c0c0c0}                                & \cellcolor[HTML]{c0c0c0}                       & \cellcolor[HTML]{c0c0c0}                              \\
\multirow{-2}{*}{\cellcolor[HTML]{c0c0c0}GADGET} & \multirow{-2}{*}{\cellcolor[HTML]{c0c0c0}0.59$\pm$0.08} & \multirow{-2}{*}{\cellcolor[HTML]{c0c0c0}\textbf{8.2}} & \multirow{-2}{*}{\cellcolor[HTML]{c0c0c0}\textbf{95.5}} & \multirow{-2}{*}{\cellcolor[HTML]{c0c0c0}8.3$\pm$2.2} & \multirow{-2}{*}{\cellcolor[HTML]{c0c0c0}0.61$\pm$0.07} & \multirow{-2}{*}{\cellcolor[HTML]{c0c0c0}\textbf{7.1}} & \multirow{-2}{*}{\cellcolor[HTML]{c0c0c0}\textbf{97.7}} & \multirow{-2}{*}{\cellcolor[HTML]{c0c0c0}9.8$\pm$2.8} & \multirow{-2}{*}{\cellcolor[HTML]{c0c0c0}0.71$\pm$0.17} & \multirow{-2}{*}{\cellcolor[HTML]{c0c0c0}\textbf{6.7}} & \multirow{-2}{*}{\cellcolor[HTML]{c0c0c0}\textbf{97.9}} & \multirow{-2}{*}{\cellcolor[HTML]{c0c0c0}8.7$\pm$2.4} & \multirow{-2}{*}{\cellcolor[HTML]{c0c0c0}0.67$\pm$0.09} & \multirow{-2}{*}{\cellcolor[HTML]{c0c0c0}\textbf{11.4}} & \multirow{-2}{*}{\cellcolor[HTML]{c0c0c0}\textbf{89.7}} & \multirow{-2}{*}{\cellcolor[HTML]{c0c0c0}8.2$\pm$2.0} \\ \hline
\end{tabular}
\end{adjustbox}
\end{table}

\begin{figure}[t]
\centering
\includegraphics[width=1\columnwidth]{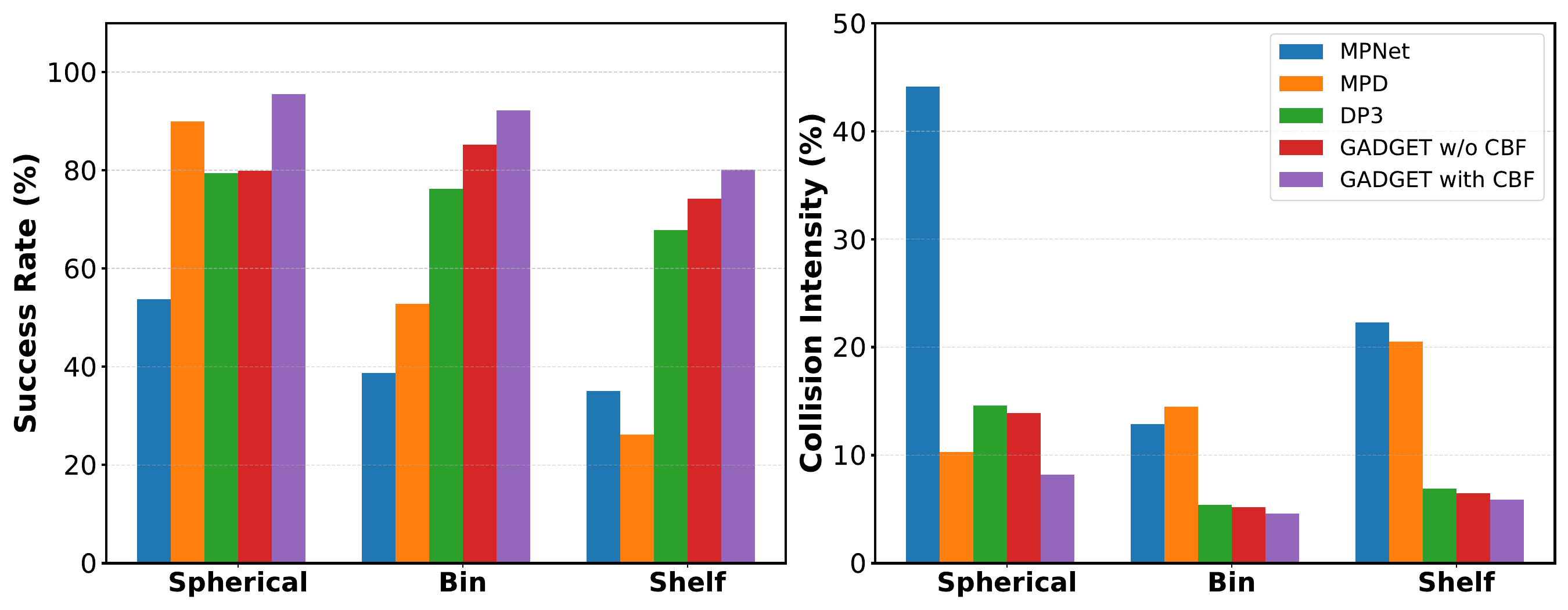}
\caption{Success rate (left) and collision intensity (right) across spherical, bin, and shelf environments. GADGET outperforms learning-based baselines (MPNet, MPD, DP3), achieving higher success with lower collision rates, while CBF guidance further enhances safety.}
\label{fig:barchart}
\end{figure}

Tables~\ref{benchmarking} and~\ref{tab:robot-transfer} highlight the effectiveness of GADGET across both environment and robot generalization benchmarks (It is worth mentioning that the system used for training and testing has a 3.500 GHz AMD Ryzen 9 processor with 64 GB RAM
and NVIDIA 3090 GPU). We used 1000 different scenarios (random start/goal configurations and random obstacle positions) in each environment. Although GADGET is being trained just on randomized spherical environment, it demonstrates zero-shot generalization to previously unseen test scenarios. In the evaluation, the position of obstacles are randomized and the test scenes are visually and geometrically distinct from those encountered during training (Bin and shelf). GADGET succes rate consistently remains high with low collision intensity (Figure~\ref{fig:barchart}). This confirms that GADGET generalizes effectively to novel environments without retraining (a clear answer to \textbf{Q1}). 

To demonstrate the performance of GADGET with different previously unseen robots, we deployed the model on manipulators with distinct kinematics such as Franka Panda, Gen3 (7DoF/6DoF), and UR5. GADGET sustains high SR and low CI (Cross-robot transfer \textbf{(\textbf{Q2})}). This transfer is driven primarily by the CBF-guided diffusion, which imposes geometry-aware safety constraints at inference time using the forward kinematics of the deployed robot, thereby providing corrective gradients that adapt to variations in link lengths, joint limits, and Jacobians. While the diffusion model learns joint-space priors implicitly conditioned on the training robot's morphology (Franka Panda), the CBF layer acts as a morphology-specific adapter that steers trajectories toward collision-free regions for the new embodiment. In contrast, MPD, DP3, and MPNet which lack such runtime geometric feedback degrade more noticeably when transferred to unseen robots. We note that this transferability is most effective among manipulators with similar kinematic topologies (serial chains with comparable DOFs); transfer to radically different morphologies (e.g., parallel robots, soft manipulators) remains an open challenge. 

To demonstrate the effect of CBF guidance, we compared the performance of GADGET with and without this guidance signal. The results show that the addition of safety constraints markedly reduces collision intensity and raises success rates, particularly in bin-picking and shelf-manipulation environments where narrow passages often cause other planners to fail (\textbf{Q3}). This indicates that CBF-guided guidance improves the safety and enhances reliability in challenging environments.

Finally, in benchmarking against state-of-the-art planners (\textbf{Q4}), GADGET demonstrates a favorable performance in all of the metrics. Classical sampling-based methods such as Bi-RRT and BIT* achieve high success rates, but only under generous search budgets. Specifically, we cap BIT* at 1 second to match our real-time budget; this naturally limits its optimality compared to longer runs. Moreover, Bi-RRT tends to produce excessively long trajectories, since it grows trees locally and connects samples opportunistically, often resulting in detours that are not globally efficient. In contrast, GADGET conditions on the full set of occupied voxel coordinates representing the scene and learns global spatial priors, enabling it to generate more direct and corridor-following paths. Interestingly, although GADGET is trained on Bi-RRT demonstrations, its diffusion-based denoising acts as a form of implicit shortcutting, capturing common structural patterns from many examples and often producing trajectories that are smoother and shorter than the teacher itself. Unlike MPNet, MPD, and DP3 which suffer from limited generalization and dependence on demonstrations, GADGET achieves consistently higher success rates, lower collision intensity, and competitive planning times across diverse environments and robot platforms, all while maintaining path lengths comparable to existing methods.

\subsection{Design Principles for Generalization}

The primary goal of GADGET is to achieve zero-shot generalization to out-of-distribution (OOD) environments and unseen robot embodiments without the need for fine-tuning. This capability is crucial for real-world deployment, where robots often encounter novel workspace layouts, obstacle geometries, and kinematic structures. GADGET attains this generalization by combining a geometry-centric scene encoding, end-to-end conditional training, and inference-time safety shaping through CBFs.

\begin{figure}[t]
\centering
\includegraphics[width=0.5\columnwidth]{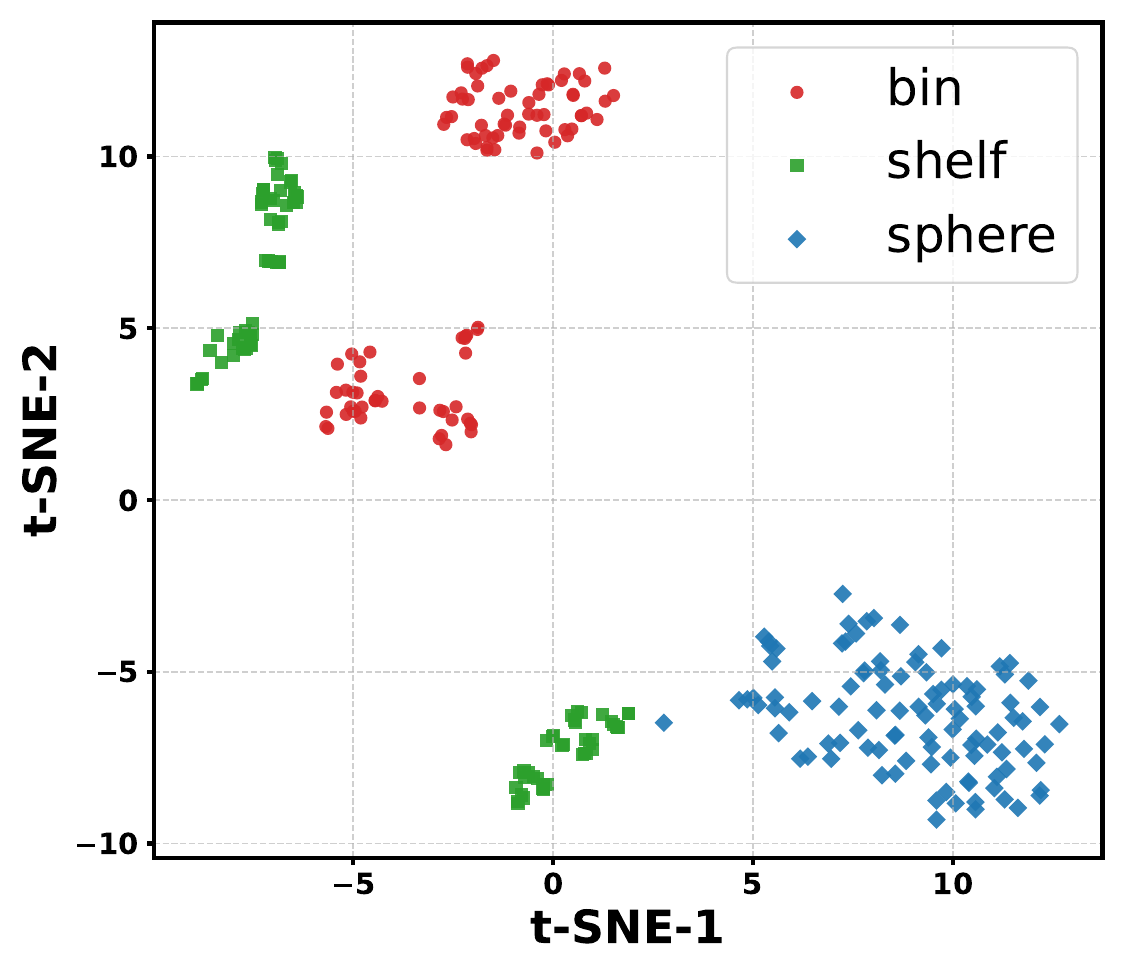}
\caption{t-SNE visualization of scene embeddings for spherical, bin, and shelf environments. The embeddings form distinct clusters, confirming that the latent distributions differ across environment types while still supporting generalization in planning.}
\label{fig:tsne}
\end{figure}

\paragraph{Geometry-Centric Scene Representation.}  
GADGET employs a sparse voxel encoding that is a product of multi-view voxel carving to reperesent the workspace. The planner takes as input the coordinates of occupied voxels. The representation although only encodes occupied areas but indirectly informs about the free space's structure through geometric context. Thus, the diffusion model is made capable of reasoning about geometry in a category-agnostic manner, which means it can be deployed in shelves or bins, for instance, that were never seen during the training phase. The generated scene embeddings (Figure~\ref{fig:tsne}) show clear latent clusters corresponding to different workspace types (spherical, bin, shelf). Despite these differences, the planner generalizes well, indicating that the diffusion model leverages geometric structure rather than memorizing specific environment categories.

\paragraph{End-to-End Conditional Training.}  
The occupied voxel coordinates $s$ are mapped into a latent embedding $z = \phi_\psi(s)$ by a scene encoder $\phi_\psi$. This embedding, together with the start and goal joint configurations $(q_s, q_g)$, forms the conditioning input $y = \{q_s, q_g, \phi_\psi(s)\}$ for the denoising diffusion model $\epsilon_\theta$. During training, the model acquires the skill of predicting and removing noise from the trajectories $\tau_t$ following the standard diffusion objective:
\begin{equation}
    \mathcal{L}(\theta, \psi) = 
    \mathbb{E}_{\tau_0 \sim \mathcal{D}, \, t \sim \mathcal{U}(1,N), \, \epsilon \sim \mathcal{N}(0,I)} 
    \Big[ \, \| \epsilon - \epsilon_\theta(\tau_t, t, y) \|^2 \, \Big],
\end{equation}
where $\tau_0$ is an expert trajectory from the dataset and $\tau_t$ is its noisy version at diffusion step $t$. Importantly, there are no direct penalties for collisions that are applied in the training phase. Instead, the encoder and denoiser are jointly optimized end-to-end purely through this denoising objective, which pressures the encoder to produce geometry-aware, semantically meaningful embeddings that remain consistent across environments.

\paragraph{Inference-Time Safety Guidance.}  
At inference, GADGET denoises a randomly initialized trajectory over $N$ steps while applying two complementary forms of guidance. First, classifier-free guidance sharpens the conditional distribution on $(q_s, q_g, \phi_\psi(s))$, improving the model’s sensitivity to task and scene context. Second, a differentiable \emph{CBF-inspired loss} $\mathcal{L}_{\text{CBF}}$ is computed dynamically at each step based on the robot’s forward kinematics and the voxel-derived scene structure. The gradient of this loss biases the reverse diffusion update away from obstacles (Eq.~\ref{eq:final_updat}), steering the trajectory toward safety-compliant regions of the state space. Although the training objective is agnostic to safety constraints, this energy-based safety shaping at inference significantly enhances robustness and feasibility in unseen environments.

\paragraph{Robot-Adaptable Formulation.}
GADGET’s generalization ability across robots relies on its separation of learned joint-space trajectory priors from the robot-specific geometric constraints enforced during inference. While the diffusion model learns collision-free joint-space distributions grounded in the kinematic structure of the training robot (e.g., Franka Panda), it does not explicitly encode robot kinematics. Instead, at inference time, the robot-dependent forward kinematics are incorporated within the CBF loss to compute link-to-obstacle distances and gradients, enabling geometry-aware safety corrections. This modular design allows the same trained diffusion model to be adapted online to new robot embodiments without retraining, provided the new morphology shares fundamental kinematic characteristics (e.g., serial chains with comparable DOFs). However, as the learned distribution reflects the morphology of the training robot, the transfer is not universal: trajectories suitable for one kinematic chain may be suboptimal or infeasible for radically different robots. Thus, the CBF layer serves as a vital morphology-specific safety interface, steering trajectories for safe cross-embodiment transfer within a constrained family of manipulators.
 
The combination of these principles provides an explanation for GADGET's success in non-training environments. The encoder–denoiser duo, tuned through end-to-end optimization, acquires a latent coordinate system associated with geometry instead of learned layouts. Although unseen environments produce different latent distributions that are disjoint from those seen in training, their codes remain interpretable by the denoiser because the semantics of the latent space are consistent. Additionally, inference-time safety guidance reinforces feasibility further that makes zero-shot planning across new environments and robots possible.

\subsection{Ablation on CBF Guidance Parameters}
\label{sec:ablation_cbf}

To determine the influence of the CBF parameters (the safety margin $\epsilon$ and the temporal constraint factor $\alpha_{\text{cbf}}$) on the planner's behavior, an ablation study is conducted in the spherical environment. The safety margin $\epsilon$ defines the minimum required distance from the obstacles. The increasing of $\epsilon$ leads to a more conservative safety buffer, thereby reducing the risk of collision but at the same time prolonging the trajectory as the planner has to go further away from the obstacles. The temporal factor $\alpha_{\text{cbf}}$ determines the aggressiveness of the planner in moving towards the safety boundary between the denoising steps. Small $\alpha_{\text{cbf}}$ values demand careful, slow updates, while large $\alpha_{\text{cbf}}$ values relax the constraint, which can lead to faster convergence but also overshooting during the passage near obstacles.

The impact of these parameters on the safety–performance was assessed by analyzing four main metrics. The success rate (SR) reflects the portion of queries that managed to attain the goal without any collisions and higher values indicate more reliable performance. The PL captures the total joint-space distance of the planned trajectory. The minimum clearance ($d_{\min}$) indicates the closest distance between any robot link and the nearest obstacle along the trajectory, with larger values indicating safer motion; this metric naturally grows as $\epsilon$ increases due to the enforced wider safety margin. Finally, CI represents the fraction of trajectory waypoints in collision, where lower values reflect stricter adherence to hard safety constraints. It should be acknowledged that the occasional invasion of the soft safety buffer (determined by $\epsilon$) does not invariably signify unsafe motion since the planner might still steer clear of collisions while taking advantage of the broader feasible region to enhance success rate.

\begin{figure}[t]
  \centering
  \includegraphics[width=\linewidth]{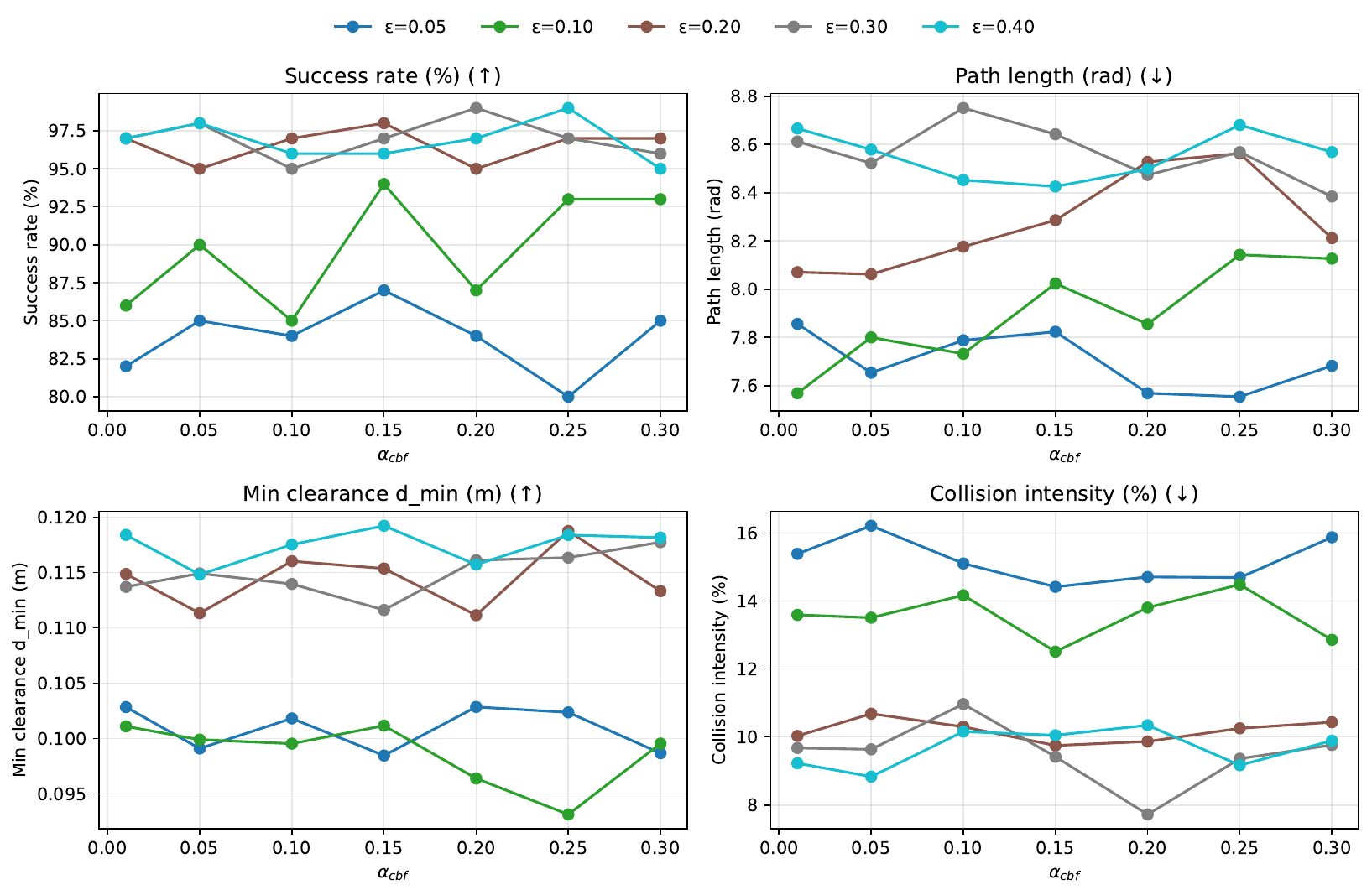}
  \caption{\textbf{Ablation on CBF guidance parameters.}
  Each curve shows the effect of $\alpha_{\text{cbf}}$ for a fixed safety margin $\epsilon$.
  \textbf{(Top-left)} Success rate (SR): moderate $\alpha_{\text{cbf}}$ (typically $0.10$–$0.20$) yields the best feasibility; very small $\alpha_{\text{cbf}}$ can overconstrain updates and very large $\alpha_{\text{cbf}}$ can destabilize denoising near obstacles.
  \textbf{(Top-right)} Path length (PL): increases with $\epsilon$ because larger margins shrink the feasible space and force detours, illustrating an SR–PL trade-off.
  \textbf{(Bottom-left)} Minimum clearance ($d_\mathrm{min}$): higher is better; confirms that trajectories remain collision-free even when soft margins are occasionally entered.
  \textbf{(Bottom-right)} Collision intensity (CI): lower is better; near-zero CI with high SR indicates robust safety in practice.}
  \label{ablation_cbf_sr_pl_clearance_ci}
\end{figure}

The impact of the CBF guidance parameters $\alpha_{\text{cbf}}$ and $\epsilon$ on performance and safety is shown in Figure \ref{ablation_cbf_sr_pl_clearance_ci}. The success rate shows a characteristic inverted–U dependence on $\alpha_{\text{cbf}}$. Very small $\alpha_{\text{cbf}}$ values overly constrain updates and reduce feasibility, whereas very large values relax safety constraints too aggressively and can destabilize the denoising process near obstacles. Apart from that, SR most of the time increases as $\epsilon$ gets larger, mainly because the planner benefits a wider feasible region and stronger safety constraints.

The increment in path length that comes with the increase in $\epsilon$ is due to the reduction of the free configuration space as a result of wider safety margins and the necessity of detours around the obstacles. The implication is that there is a fundamental trade-off; the larger the value of $\epsilon$, the better the SR and the safer the operation but at the cost of longer and less efficient trajectories. However, this trade-off can be adjusted according to the application. For example, in the case of a safety-critical task, we can give priority to the success rate or, if the task is considered to be very efficient, we can prefer shorter trajectories. The average minimum clearance $d_{\min}$ is a direct indicator of how much the robot approaches the obstacles. As it was expected, $d_{\min}$ is larger with the larger $\epsilon$, thus it proves that the planner keeps larger clearances when this is necessary. However, at the same time, the collision intensity remains near zero for most parameters. This implies that even when the trajectories go into the soft safety zone, they do not violate the hard collision constraints very often. These patterns confirm the strength and flexibility of GADGET over a very wide range of parameter settings and without the need for intensive fine-tuning.


\begin{figure}[t]
\centering
\includegraphics[width=1\columnwidth, trim=0 50 0 30, clip]{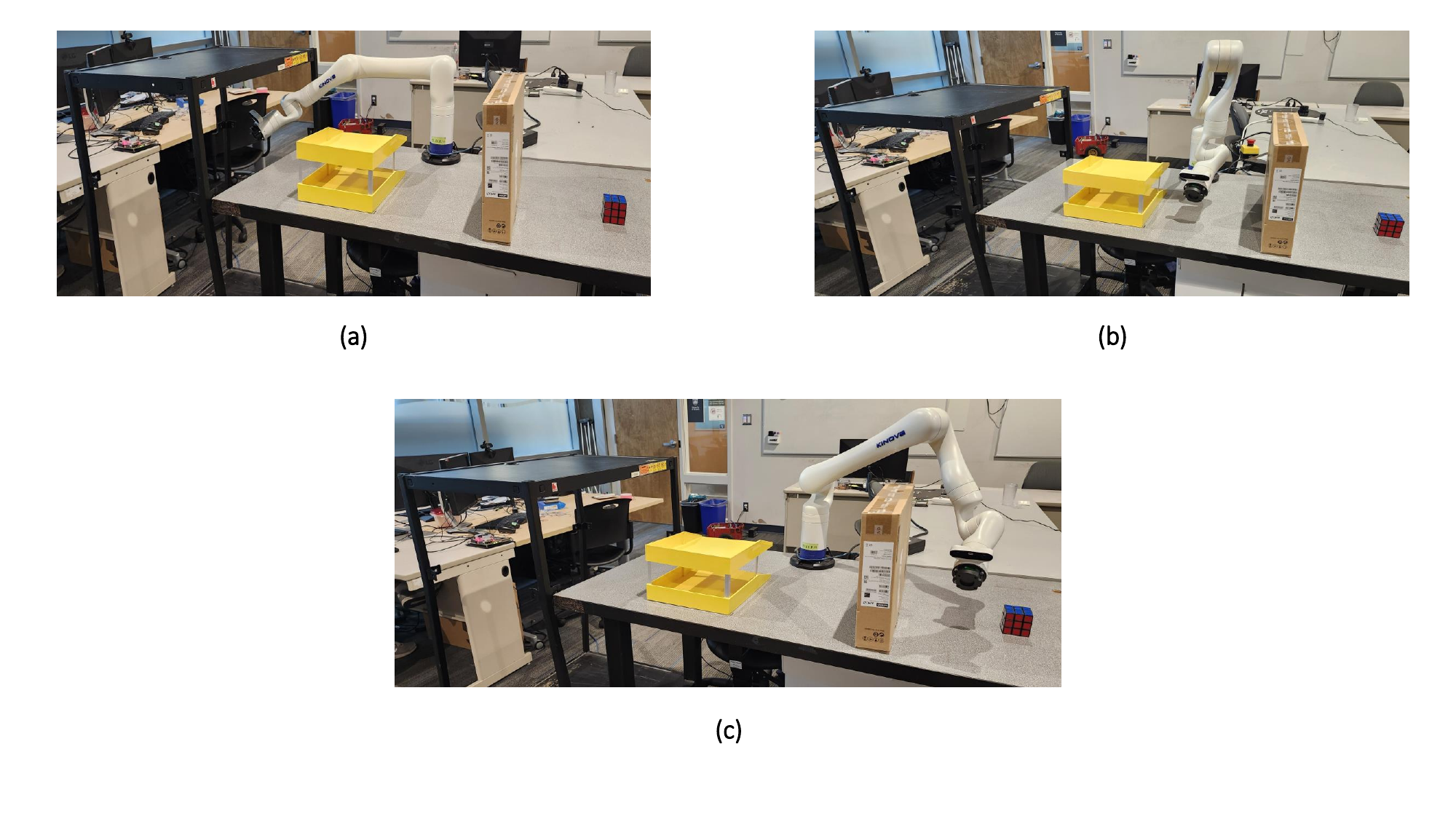}
\caption{Practical implementation setup on the Kinova Gen3. The robot starts from configuration (a), moves to an intermediate goal (b), then proceeds to target (c), and finally will be back to (a). GADGET is tasked with generating collision-free trajectories between these waypoints.}
\label{realworld}
\end{figure}

\section{Real-world Implementation} \label{real}

\begin{figure}[t]
\centering
\includegraphics[width=1\columnwidth, trim=20 120 20 70, clip]{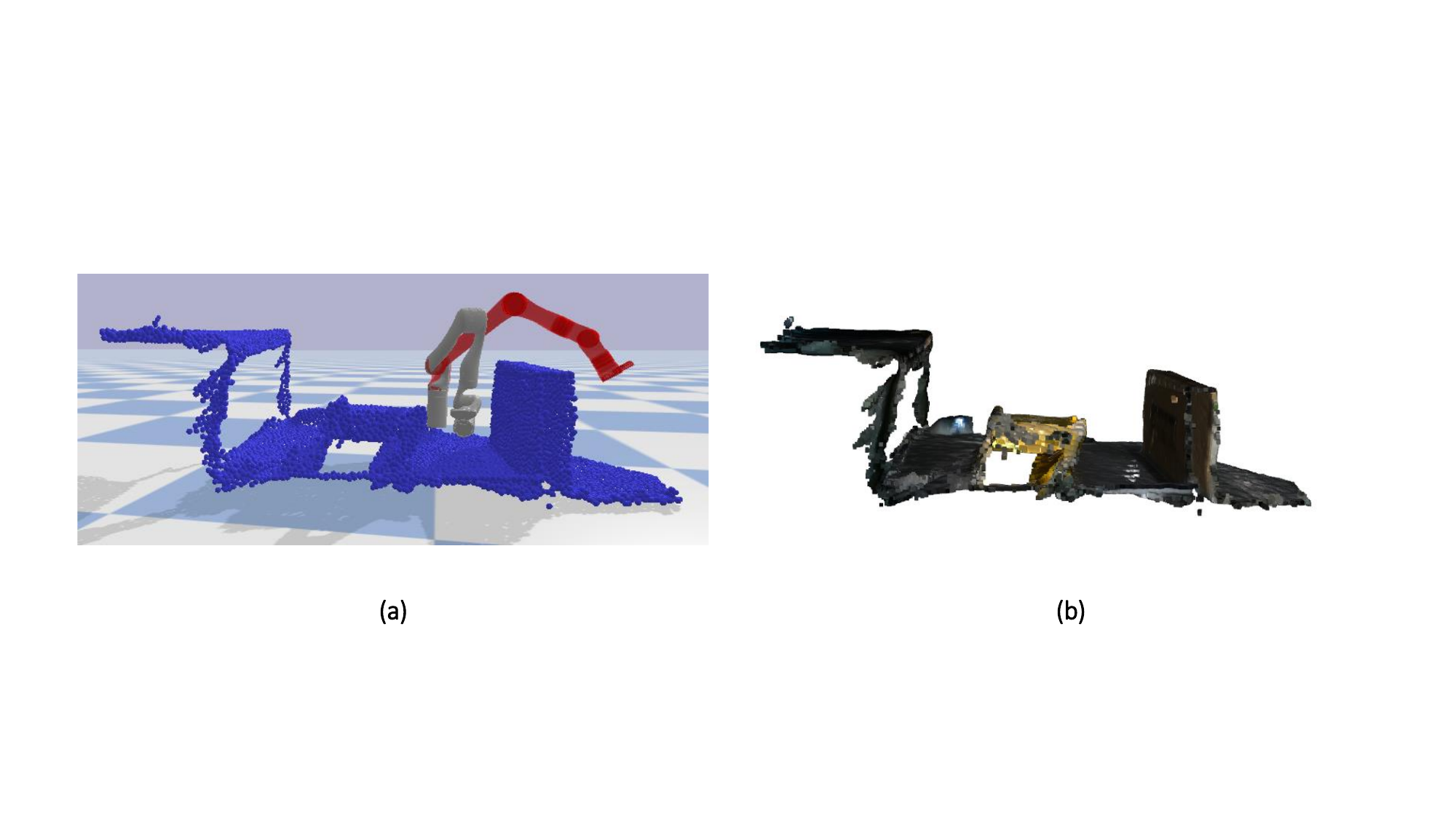}
\caption{(a) Digital twin of the environment in PyBullet used for planning and collision checking. (b) 3D reconstruction of the same environment obtained from the depth camera.}
\vspace{-2em}
\label{3D}
\end{figure}

The real-world execution of GADGET follows a five-stage pipeline:

\begin{enumerate}
    \item \textbf{Environment Scanning:} An end-effector-mounted RGB-D camera captures multiple views of the workspace from different angles. These observations provide depth data critical for scene understanding.

    \item \textbf{3D Reconstruction:} The captured multi-view depth images are fused using open3D library to reconstruct a 3D model of the workspace. 

    \item \textbf{Digital Twin Generation in PyBullet:} To generate  a digital twin of the real workspace the reconstructed voxel map is imported into PyBullet (Figure~\ref{3D}). This simulated replica will be used for collision evaluation.

    \item \textbf{Trajectory Generation via GADGET:} Given the voxel-encoded scene and goal condition, GADGET smaples a set of joint-space trajectory using its conditional diffusion model, guided at inference time by CBF-based safety constraints.

    \item \textbf{Robot Execution:} The generated trajectory is streamed to the physical robot controller for execution. Since planning occurs directly in joint space, no additional inverse kinematics is required.
\end{enumerate}

This pipeline is used to demonstrate GADGET’s ability to real-world scenarios without any retraining. By combining scene perception with CBF-guided generative planning, the framework acts as a robust zero-shot path planner, capable of adapting to arbitrary changes in workspace geometry. Figure~\ref{realworld} shows our real-world setup. We sampled 50 trajectories from GADGET for this task which 38 of them were successful. We implemented those on our real robot using the kortex API~\cite{kinova}. We obtained 30/38 collision-free paths, demonstrating GADGET's efficacy of
generating diverse success solutions. We hypothesize that the
remaining 8 trajectories that collided with the obstacles
are due to the inaccurate 3D reconstruction and digital twin generation.

\begin{figure}[t]
\centering
\includegraphics[width=1\columnwidth, trim=0 120 0 70, clip]{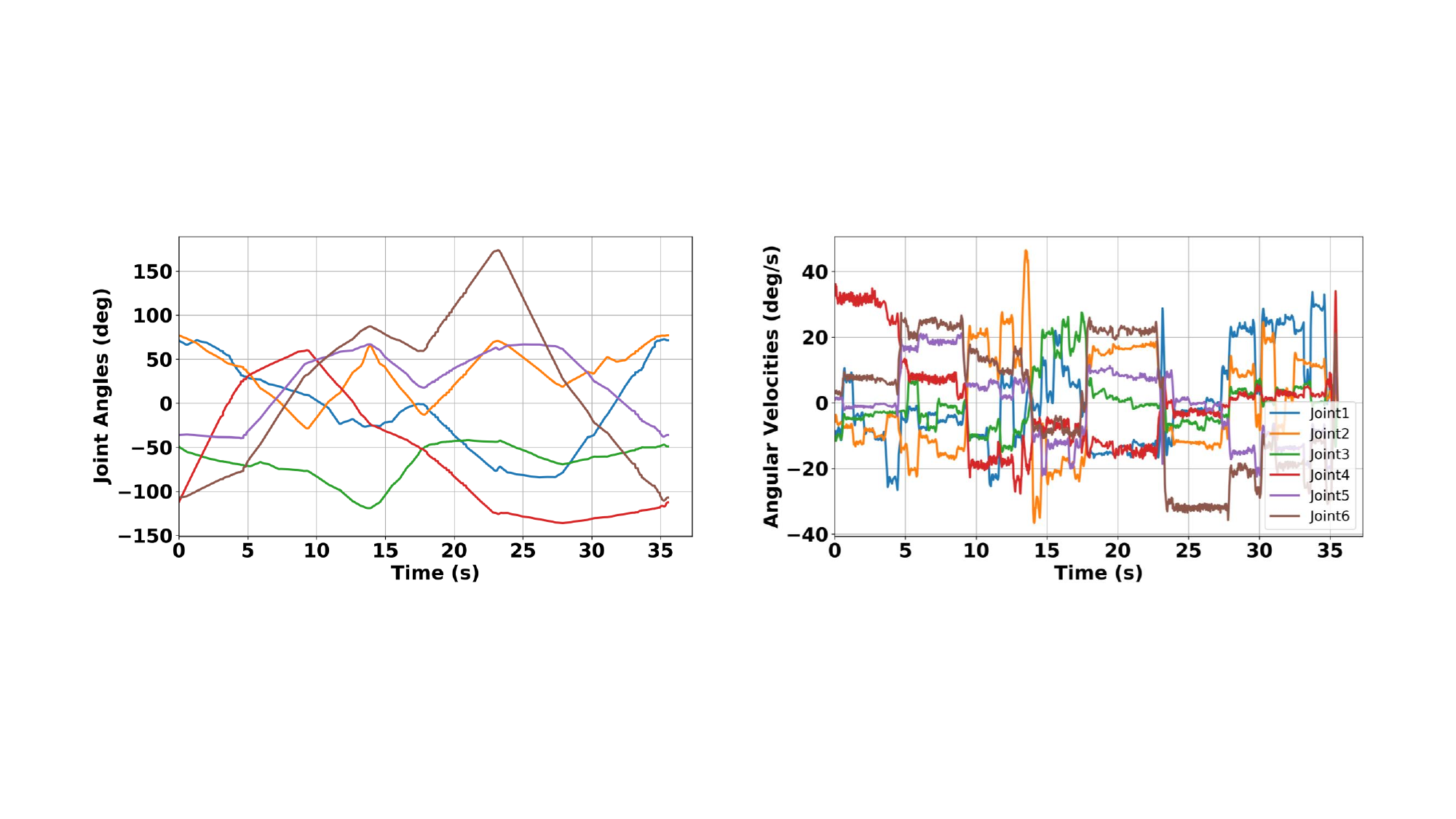}
\vspace{-2em}
\caption{Experimental dataset samples collected from the real Kinova Gen3 setup. These data are used to assess GADGET’s generalization under practical conditions.}
\label{realdata}
\end{figure}

The success of this deployment highlights the practical potential of diffusion-based planners for industrial robotic manipulation. Notably, the real-world implementation employs the same model trained entirely in simulation using spherical obstacle environments and the Franka Emika Panda arm, demonstrating strong generalization. To further illustrate the deployment, Figure~\ref{realdata} presents representative real-world execution data recorded during experiments. These results confirm that the trajectories generated by GADGET in the digital twin can be reliably executed on hardware\footnote{A video demonstrating the simulation results and real-world deployment is available at the following link: \href{https://youtu.be/6ocxqQTG2d0}{https://youtu.be/6ocxqQTG2d0}}.

\section{Conclusions} \label{con}

This paper proposed GADGET, a generalizable and safety-aware diffusion-based path planning framework for robotic manipulators. Using voxelized scene encoding and conditioning on start/goal configurations, the proposed planner generates diverse and feasible joint-space trajectories that can be adapted to previously unseen environments. A CBF--inspired guidance mechanism was integrated directly into the diffusion inference process to ensure safety. This guidance mechanism steers denoising steps away from collisions. GADGET's dual phase conditioning (combining classifier-free guidance and classifier-guided diffusion) enables generalization over unseen environments and robotic arms. Experimental results indicate that GADGET consistently achieves high success rates, low collision intensity, and comparable total path lengths across diverse environments, including spherical obstacles, bin-picking, and shelf manipulation. Moreover, CBF-guided corrections plus the robot-adaptable joint-space formulation enabled zero-shot transfer across manipulators with distinct kinematics, such as Franka Emika Panda, Kinova Gen3, and UR5. Additionally, to validate the practicality of the framework in real-world settings, it was deployed on the Kinova Gen3. In summary, GADGET establishes diffusion-based generative planning as a promising paradigm for robotic manipulation, unifying geometric scene understanding, task conditioning, and safety-aware inference. GADGET's ability to generate collision-free trajectories across unseen environments and robot embodiments without retraining opens new pathways for scalable, deployment-ready motion planning in manufacturing where adaptability to dynamic layouts and diverse hardware is essential for real-world viability.
Future work will include extending this framework toward handling dynamic environments by integrating richer sensor modalities and reducing inference time to support real-time adaptive manipulation in industrial and collaborative robotics.



\printcredits
\section*{Declaration of competing interest}
The authors declare that they have no known competing financial interests or personal relationships that could have appeared to influence the work reported in this paper.

\section*{Acknowledgment}
This work is the results of the research project funded by Mathematics of Information Technology and Complex Systems (MITACS) and Apera AI under IT16412 Mitacs Accelerate.

\section*{Data Availability}
Data will be made available on request.

\bibliographystyle{unsrtnat}  

\bibliography{cas-refs}





\end{document}